\documentclass{article}
\usepackage{iclr2027_conference,times}
\usepackage{amsmath,amssymb,amsthm}
\usepackage{graphicx,booktabs,multirow,tabularx,array}
\usepackage{placeins,enumitem,wrapfig,capt-of}
\usepackage{hyperref,xurl}
\graphicspath{{pics/}}
\hypersetup{hidelinks,hidelinks,hidelinks,pdfauthor={},pdfcreator={},pdfproducer={}}
\let\cite\citep

\iclrfinalcopy
\newcommand{\CaFEAuthorBlock}{%
\textbf{Hongyin Zhang$^{1}$}\thanks{Equal contribution.}, \textbf{Mingyi Xiang$^{1}$}\footnotemark[1], \textbf{Rongzhao Chen$^{2}$}, \textbf{Yankai Wang$^{1}$}, \textbf{Jialin Qiao$^{2}$},\\
\textbf{Xiangdong Huang$^{1}$}, \textbf{Xiaoou Ding$^{3}$}, \textbf{Chen Wang$^{1}$}, \textbf{Zhongyi Pei$^{1,\boxtimes}$}, \textbf{Jianmin Wang$^{1}$}\\
\textnormal{\small$^{1}$School of Software, BNRist, Tsinghua University}\\
\textnormal{\small$^{2}$Timecho Ltd, Beijing, China}\\
\textnormal{\small$^{3}$Harbin Institute of Technology}\\
\textnormal{\small\texttt{\{hy-zhang25, xiangmy21\}@mails.tsinghua.edu.cn}}\\
\textnormal{\small$\boxtimes$~\texttt{peizhyi@tsinghua.edu.cn}}%
}

\title{Beyond Static Leaderboards: Capability-Focused Evaluation with Structural Diversity for Time Series Foundation Models}
\author{\CaFEAuthorBlock}
\begin{document}
\maketitle
\lhead{}
\begin{abstract}
Static public benchmarks for time series foundation models (TSFMs) typically have fixed and narrow test distributions. Their repeated reuse encourages benchmark-specific tuning, making reported performance reflect adaptation to the fixed test set rather than genuine forecasting generalization. In addition, a single aggregate score cannot fully characterize model capabilities across distinct temporal structures and input relationships. We propose CaFE (Capability-Focused Evaluation), a refreshable evaluation extension grounded in original benchmark data that enriches static benchmarks with structural diversity and enriches single-score black-box evaluation with multi-dimensional, capability-oriented assessment. CaFE first profiles the original benchmark and defines eight Structural Features (SFs) that characterize its temporal structure. For each SF, it resamples the original data according to its structural support, removes edge-of-distribution samples, and augments the retained samples by varying the SF's salience, producing samples with different structural profiles. After filtering out out-of-distribution samples while preserving diversity, CaFE yields a refreshed evaluation set. TSFMs are then evaluated on this set using both conventional forecast-error metrics and intervention-response metrics, giving each model a per-feature capability profile alongside endpoint accuracy. Experiments on GIFT-Eval and FEV-Mini20 show that CaFE yields stable evaluation results and reveals per-feature strengths and weaknesses that a single aggregate score hides. Moreover, on genuine data filtered by structural condition, models selected by CaFE outperform those chosen by the aggregate leaderboard. CaFE therefore makes model selection conditional on the structural response of interest.
\end{abstract}

\section{Introduction}
\label{sec:intro}

Time series foundation models (TSFMs), pretrained on large and diverse data corpora, are increasingly used for forecasting across domains such as energy, transportation, finance, and cloud operations~\cite{chronos,timesfm,moirai}. As more TSFMs become publicly available, selecting the right model for a specific forecasting task has become a critical practical challenge. Existing benchmarks evaluate TSFMs on widely reused datasets such as ETT, Electricity, and Traffic~\cite{tsfmbench,gifteval,zhou2021informer,trindade2015electricity,lai2018lstnet}, typically summarizing performance with a single aggregate score. These leaderboards rank models by overall performance and have become the de facto standard for TSFM comparison.

\begin{wrapfigure}{R}{0.46\textwidth}
  \vspace{-6pt}
  \centering
  \includegraphics[width=0.46\textwidth]{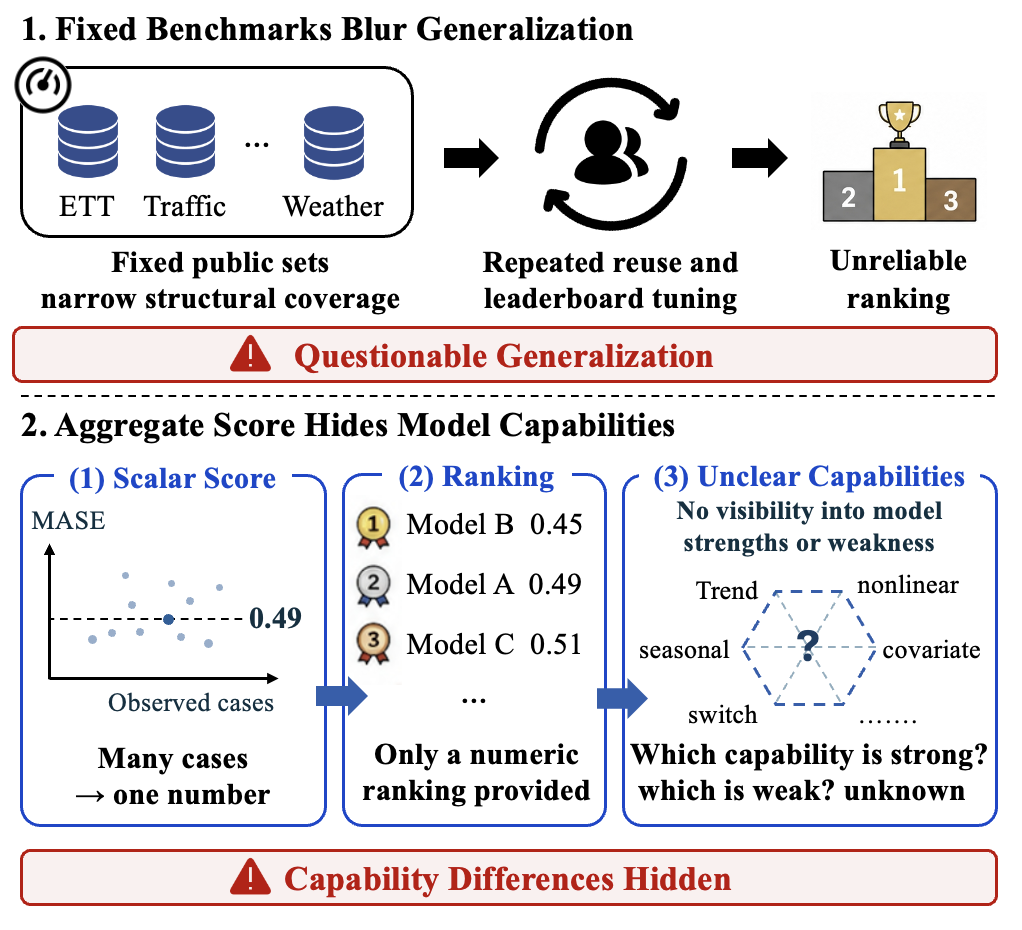}
  \small
  \caption{A static test set provides one finite snapshot of structural conditions, while a single aggregate score obscures how forecasts respond across those conditions.}
  \vspace{-12pt}
  \label{fig:motivation}
\end{wrapfigure}

Fixed public test sets create two intertwined problems. First, their repeated reuse conflates forecasting ability with familiarity with the specific test content, and their narrow structural coverage leaves structural diversity underrepresented. Second, a single aggregate score collapses many distinct observed cases into one ranking. As a result, the model that attains the best endpoint accuracy may differ from the model that responds faithfully to a controlled structural change, and existing benchmarks offer no direct way to inspect that difference (Figure~\ref{fig:motivation}).

To address these two problems, we present \textsc{CaFE} (Capability-Focused Evaluation), a refreshable evaluation extension grounded in original benchmark data. CaFE enriches static benchmarks with structural diversity and enriches single-score evaluation with multi-dimensional, capability-oriented assessment. Concretely, it first profiles the original benchmark and defines eight Structural Features (SFs) spanning intrinsic and relational structure. Table~\ref{tab:structural-features} summarizes the eight SFs and their controlled changes. Guided by these SFs, it resamples the original data along specific structural dimensions and removes edge-of-distribution samples to establish controlled structural conditions. It then augments the resampled data by varying the salience of each SF, producing samples with different structural profiles, and filters out out-of-distribution augmented samples while preserving diversity, yielding a refreshed evaluation set. By spanning a wider range of structural conditions than the fixed test set, this refreshed set lowers the risk that reported performance reflects adaptation to a narrow distribution rather than genuine forecasting generalization.

On the evaluation side, CaFE measures each model along two axes. The first retains conventional forecast-error metrics, capturing endpoint accuracy on the refreshed set, as in existing benchmarks. The second introduces intervention-response metrics: for each Structural Feature, the evaluator varies the feature's salience and measures how the model's forecast changes in response, yielding a per-feature capability profile. The profile reveals structural sensitivity that endpoint accuracy alone misses: a model can forecast accurately while ignoring cross-series relationships or covariate conditions. CaFE therefore reports endpoint accuracy alongside a per-feature capability profile, exposing condition-specific strengths and weaknesses that a single aggregate score cannot express.
In summary, we make the following main contributions:
\par\smallskip\noindent\textbullet\enspace We propose \textsc{CaFE}, a refreshable evaluation extension grounded in original benchmark data that enriches static benchmarks with structural diversity and enriches single-score evaluation with multi-dimensional, capability-oriented assessment.
\par\smallskip\noindent\textbullet\enspace We introduce Structural Features (SFs) and, guided by them, resample and augment the original data under controlled structural conditions, filtering out out-of-distribution samples to yield a refreshed evaluation set. We then expose per-feature response profiles through intervention-response metrics alongside conventional forecast-error metrics, adding a multi-dimensional, capability-oriented axis to evaluation.
\par\smallskip\noindent\textbullet\enspace Across GIFT-Eval and FEV-Mini20, we show that CaFE's per-feature response profiles expose condition-specific strengths and weaknesses a single score hides, that repeated regenerations preserve the overall ranking while refreshing local winners, that training recipes can trade endpoint accuracy against response fidelity, and that models selected by CaFE outperform the aggregate leaderboard on genuine data filtered by structural condition.

\section{Related Work}
\label{sec:related}

\begin{wrapfigure}{R}{0.46\textwidth}
  \vspace{-8pt}
  \centering
  \captionof{table}{Benchmark families compared on four design properties: anchored to original instances, explicit structure, paired change, and refresh. CaFE is the only family that combines all four.}
  \label{tab:benchmark-comparison}
  \scriptsize
  \setlength{\tabcolsep}{1.5pt}
  \renewcommand{\arraystretch}{1.1}
  \resizebox{\linewidth}{!}{%
    \begin{tabular}{@{}lcccc@{}}
      \toprule
      Family & \shortstack{Original\\instances} & \shortstack{Explicit\\structure} & \shortstack{Paired\\change} & \shortstack{Refresh} \\
      \midrule
      Retrospective suites & $\checkmark$ & $\times$ & $\times$ & $\times$ \\
      TSFM suites & $\checkmark$ & $\times$ & $\times$ & $\times$ \\
      Covariate-aware suites & $\checkmark$ & $\times$ & $\times$ & $\times$ \\
      Observed-structure suites & $\checkmark$ & $\checkmark$ & $\times$ & $\times$ \\
      Live evaluation & $\checkmark$ & $\times$ & $\times$ & $\checkmark$ \\
      Synthetic mechanisms & $\times$ & $\checkmark$ & $\times$ & $\times$ \\
      \textbf{CaFE(Ours)} & $\checkmark$ & $\checkmark$ & $\checkmark$ & $\checkmark$ \\
      \bottomrule
    \end{tabular}
  }
\end{wrapfigure}

Time series foundation model evaluation spans four broad lines. Fixed retrospective and TSFM suites, including M4/M5, TFB, BasicTS+, ProbTS, TSFM-Bench, GIFT-Eval, and fev-bench, provide data, pipelines, and standardized comparisons~\cite{makridakis2020m4,makridakis2022m5,tfb,basicts,probts,tsfmbench,gifteval,shchur2025fev}. Structure-aware suites such as It's TIME and QuitoBench organize observed patterns or balanced regimes, and live platforms such as TS-Arena and Impermanent use preregistration or rolling streams~\cite{timebench,quitobench,meyer2025tsarena,impermanent}. Synthetic-generation benchmarks such as GRATIS, TSGM, TimeSynth, and SynTSBench provide programmable mechanisms and behavior or fidelity diagnostics~\cite{noisyperiodic,lowsnr,finstressts,kang2020gratis,timegan,timevae,diffusionts,ang2023tsgbench,tsgm,syntsbench,timesynthbias}. These lines vary fixed collections, observed regimes, incoming streams, or generators; the fixed suites emphasize endpoint comparisons, leaving source-task anchoring and controlled paired changes as separate design choices.

For CaFE, four properties distinguish an evaluation design: whether it anchors to original benchmark instances, makes structure explicit, applies a controlled paired change, and is refreshable. Retrospective, TSFM, and covariate-aware suites anchor to original instances but keep structure implicit and apply no paired changes; observed-structure suites make structure explicit but do not refresh; live suites refresh streams but apply no paired changes; synthetic suites expose mechanisms but abandon the source task context. CaFE combines all four, applying one Structural Feature change to an original instance under screening and then reporting endpoint error on the References and generated samples together with intervention-response fidelity. Table~\ref{tab:benchmark-comparison} summarizes this positioning.

\section{The \textsc{CaFE} Framework}
\label{sec:method}

\begin{figure}[t]
  \centering
  \includegraphics[clip,width=0.98\textwidth]{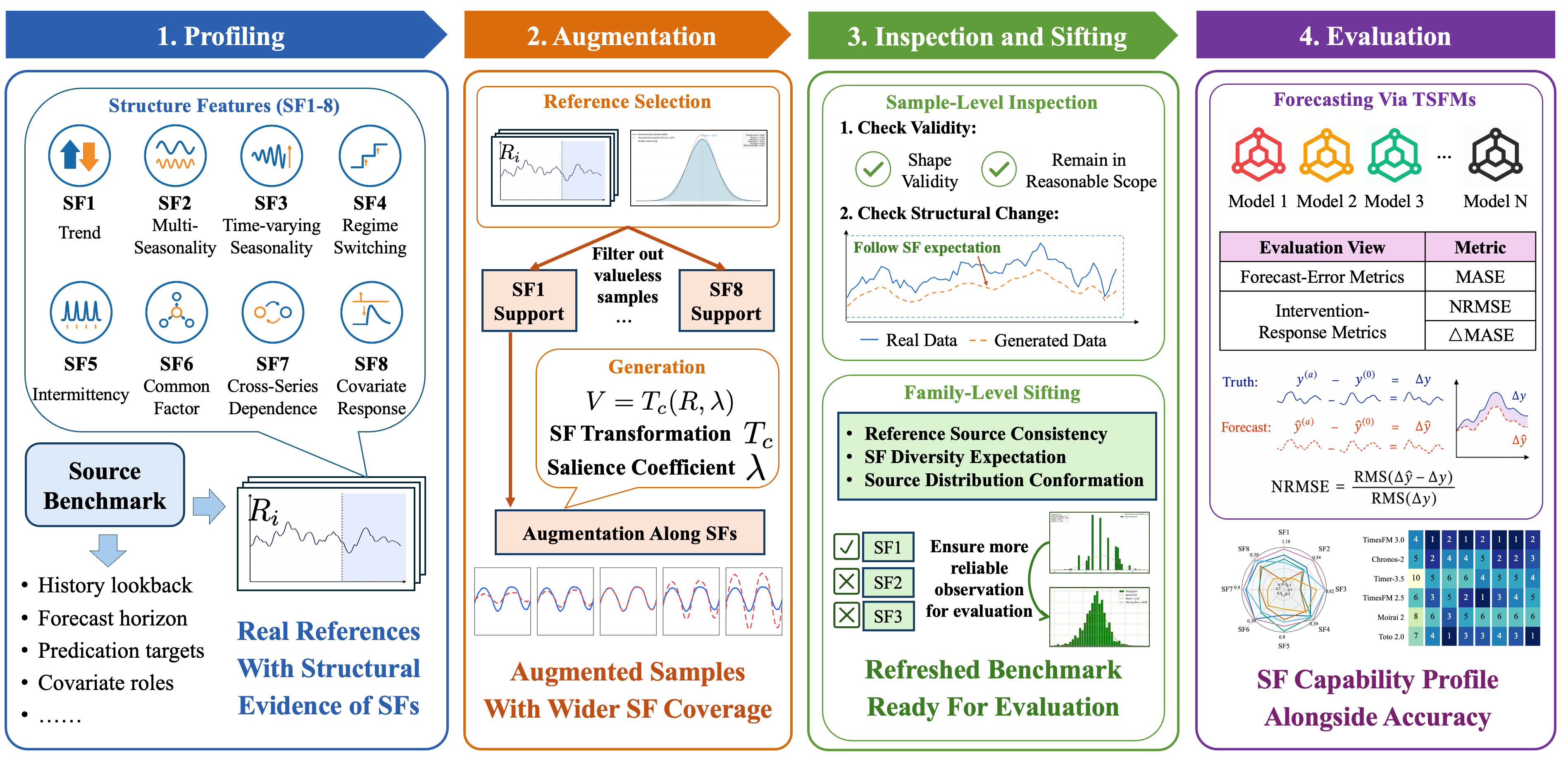}
  \caption{The \textsc{CaFE} workflow. Profiling maps source forecast windows to unchanged References; Augmentation generates, for each Reference, a family of samples at varying salience; Inspection and Sifting admit only complete families that realize the intended change; Evaluation scores each model on forecast-error and intervention-response metrics.}
  \label{fig:pipeline}
  \vspace{-4pt}
\end{figure}

\textsc{CaFE} addresses two failures of static benchmarks, a narrow structural distribution and a single aggregate score, by enriching a benchmark's own data with structural diversity and its evaluation with capability-oriented assessment. Concretely, it proceeds in three generation stages followed by an evaluation stage (Figure~\ref{fig:pipeline}). \textbf{Profiling} defines eight Structural Features (SFs) and maps each eligible source forecast window to an unchanged Reference. \textbf{Augmentation} resamples the References by structural support, removes edge-of-distribution windows, and, for each retained Reference, draws random salience coefficients to generate a family of samples spanning diverse structural conditions. \textbf{Inspection and Sifting} admits only complete families whose samples realize the intended change without drifting out of distribution. Finally, \textbf{Evaluation} scores each model along two axes.

The first axis retains forecast-error metrics, capturing endpoint accuracy on the refreshed set. The second adds intervention-response metrics that measure how faithfully a forecast tracks each induced change. Because every generated sample differs from its Reference only through a known change in one SF, the same benchmark yields both an endpoint-accuracy profile and a per-SF response profile, so model selection can follow capability for each specific SF rather than a single aggregate score.

\subsection{Profiling}
\label{subsec:profiling}
\begin{wrapfigure}{R}{0.54\textwidth}
  \centering
  \includegraphics[width=0.54\textwidth]{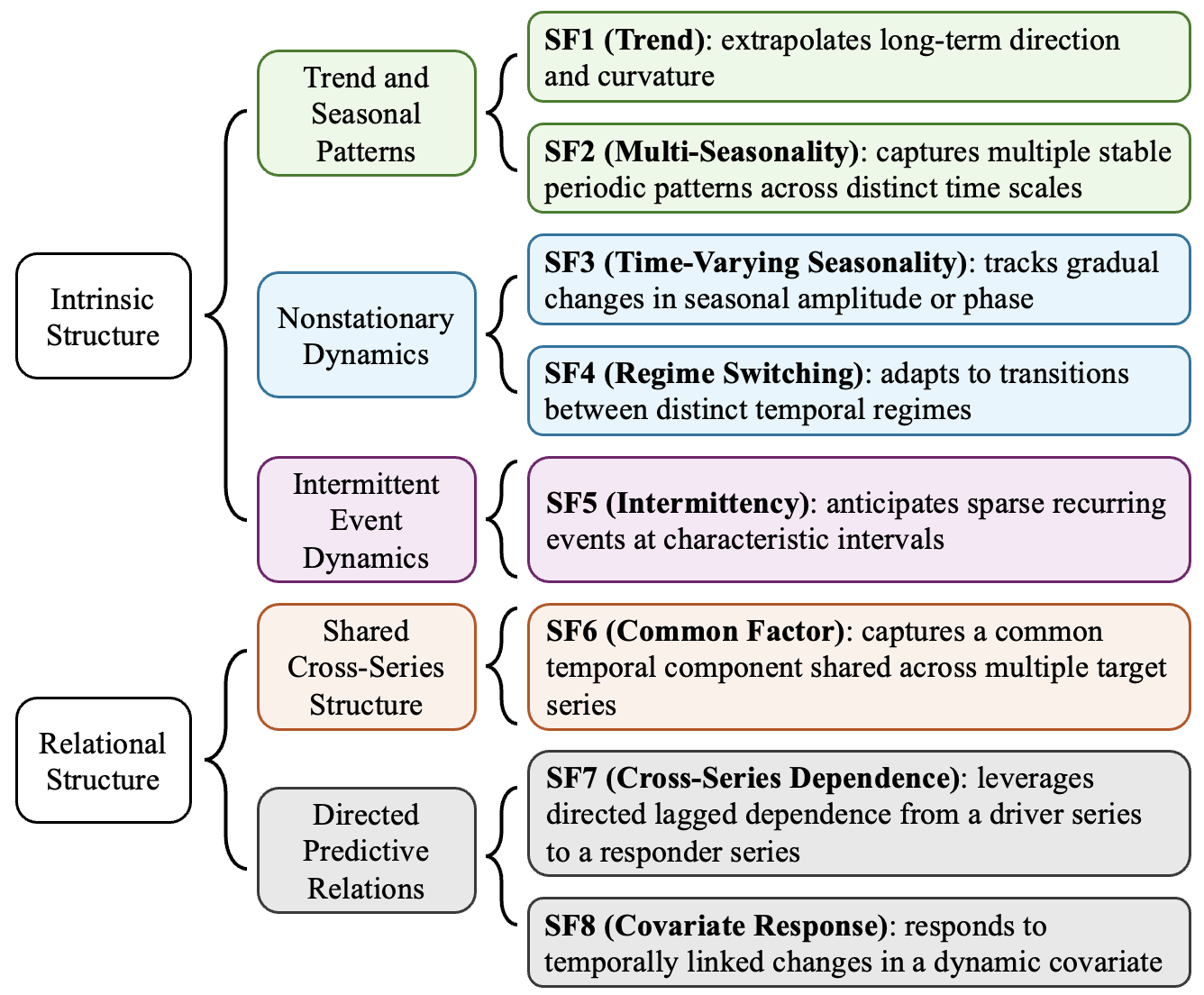}
  \caption{The Structural Features of \textsc{CaFE}: intrinsic structure for within-series patterns and relational structure for cross-series and covariate dependencies.}
  \label{fig:capability-overview}
\end{wrapfigure}

Profiling anchors every intervention to the original forecasting task. It defines eight Structural Features (SFs) and maps each eligible source forecast window to an unchanged Reference $R_i$ that preserves the targets, forecast origin and horizon, input visibility, and scoreable positions. Because the Reference is a real, unmodified forecasting example, every generated sample stays grounded in real benchmark data rather than synthetic data.
A Structural Feature is a temporal pattern or relationship whose salience can be varied in a controlled way to probe a specific forecasting capability. CaFE groups the eight features into two classes: intrinsic structure, which operates within a single target series, and relational structure, which operates across target series or between covariates and targets. Figure~\ref{fig:capability-overview} summarizes the eight features, and Appendix~\ref{app:capability-mechanisms} provides details.

\paragraph{Intrinsic Structure.} Intrinsic structure comprises five features (SF1--SF5) spanning three forms of within-series variation. Trend and Seasonal Patterns (SF1--SF2) vary in persistent direction and concurrent periodicity, testing whether forecasts extrapolate long-term structure and combine multiple temporal scales. Nonstationary Dynamics (SF3--SF4) vary how seasonal behavior and temporal regimes evolve, testing whether forecasts track changes over time. Intermittent Event Dynamics (SF5) varies the regularity of sparse recurring events, testing whether forecasts recognize predictable structure when observations are mostly inactive. Moving from persistent patterns to evolving dynamics and sparse events, these interventions distinguish different forms of temporal generalization while retaining the same task anchor.

\paragraph{Relational Structure.} Relational structure comprises three features (SF6--SF8) spanning shared and directed dependencies. Shared Cross-Series Structure (SF6) varies a temporal component common to multiple target series, testing whether forecasts recognize movement shared across targets. Directed Predictive Relations (SF7--SF8) vary a lagged relation from a driver to a responder or a target's response to a dynamic covariate, testing whether forecasts use signals from other series or covariates. Together, these features reveal whether a forecast uses information shared across targets or directed information from a specific series or covariate, while the series involved and the forecast task remain fixed. Because SF6--SF8 expose auxiliary inputs, CaFE additionally reports input utility by removing each input and measuring the resulting change in forecast error.

\subsection{Augmentation}
\label{subsec:augmentation}
Augmentation turns each Reference into a family of generated samples that differ from it only in a single Structural Feature. It first resamples the References according to their structural support for each SF, removing edge-of-distribution windows so that the refreshed set stays within the benchmark's typical structural range. For each retained Reference and selected SF, it then draws random salience coefficients, each controlling how strongly the feature is expressed, and generates one sample per coefficient. Because the coefficients are drawn over declared ranges, the resulting families span a wider and more diverse set of structural conditions than the fixed test set.

For each Reference $R$ and Structural Feature $c$, Augmentation draws a random salience coefficient $\lambda$ and generates a sample $V = T_c(R, \lambda)$. All samples generated from the same Reference and feature differ only in $\lambda$, so the family spans a range of salience levels while the forecast origin, horizon, target order, input visibility, and scoreable positions remain fixed. Each Structural Feature specifies its own transformation $T_c$, which injects the designated pattern or relation into the Reference while leaving all other structure untouched. The concrete construction of each feature and the range from which $\lambda$ is drawn are given in Appendix~\ref{app:capability-mechanisms}; for the present purpose, every generated sample is a controlled perturbation of its Reference with a known salience coefficient.

Together, the eight features and their randomly drawn salience coefficients give the refreshed set a structural coverage far wider than the fixed test set, so that reported accuracy reflects generalization across structure rather than familiarity with a narrow distribution. The generated families are then screened by Inspection and Sifting before they enter the evaluation set.

\subsection{Inspection and Sifting}
\label{subsec:inspection-sifting}
Augmentation produces candidate families, and Inspection and Sifting determine whether they can serve as evaluation data. \textbf{Inspection} works on each generated sample, verifying that the intended structural change is present, that non-target structure stays within tolerance, and that the sample remains valid and scoreable. \textbf{Sifting} works on the family linked to one Reference, requiring the family to be complete, checking that all members share the same Reference and configuration, and confirming that the family stays within the source distribution while preserving the structural diversity introduced by augmentation.

A family is admitted into the \emph{Refreshed Benchmark} only when every member passes Inspection, and the family as a whole passes Sifting. Rejected or incomplete families receive an explicit reason and do not contribute to the evaluation, while their References remain available for the original task. The checks cover basic validity, realization of the intended structural change, non-target tolerance, source distance, scoreable support, distribution difference, and diversity. See Appendix~\ref{app:protocol} for details.

\subsection{Evaluation}
\label{subsec:evaluator}
CaFE reports two metric classes. \textbf{Forecast-Error Metrics} measure endpoint accuracy on the References and the generated samples, using Reference MASE and Probe MASE. \textbf{Intervention-Response Metrics} measure whether forecasts respond to the structural change introduced by augmentation along each SF, using Forecast-Change NRMSE, the normalized root-mean-square error between the induced target change and the model's forecast change, with the Auxiliary-Input Gap ($\Delta\mathrm{MASE}$) reported additionally for SF6--SF8. Together, these metrics characterize endpoint accuracy, structural response, and auxiliary-input utility.

For a given model and Structural Feature, Forecast-Change NRMSE measures how closely the model's forecast change tracks the target change induced by varying the feature's salience. Let $\Delta y$ and $\Delta\hat y$ denote the target change and the forecast change between a generated sample and its Reference on the affected positions. Forecast-Change NRMSE is
\begin{equation}
 \operatorname{NRMSE}=\sqrt{\frac{\sum_{r=1}^{|\Delta y|}(\Delta\hat y_r-\Delta y_r)^2}{\sum_{r=1}^{|\Delta y|}(\Delta y_r)^2}},
\end{equation}
where $r$ indexes the affected positions. A value of $0$ indicates an exact response, $1$ indicates no forecast response, and lower values mean closer agreement with the induced change. Eligibility and aggregation details are given in Appendix~\ref{app:protocol}.

The Auxiliary-Input Gap measures the change in MASE after removing the designated auxiliary input, with positive values indicating that the full input helps. The two metric classes keep their own rankings: Forecast-Error Metrics summarize accuracy over References and generated samples, while Intervention-Response Metrics remain within each SF, whose controls, scales, and scoreable populations differ. Any cross-feature summary is a secondary equal-weight macro over direction-corrected feature-local ranks and does not pool raw values or define a cross-feature capability scale.

\section{Experiments}
\label{sec:experiments}
The experiments establish CaFE as an evaluation framework for time series foundation models. We first apply CaFE to GIFT-Eval (Short, Medium, and Long horizons) and fev-bench Mini-20, reporting endpoint accuracy and response fidelity for each model. On genuine data filtered by intrinsic structure, models selected by CaFE outperform those chosen by the aggregate leaderboard. We then validate the relational features SF6--SF8, verify that the augmentation is faithful, and check that the rankings are stable under regeneration. Finally, we show that the two metrics trade off under a changed training objective.

\subsection{Experimental Setup}
\label{subsec:setup}
We evaluate CaFE on three horizon subsets of GIFT-Eval~\cite{gifteval} (Short, Medium, and Long) and the Mini-20 subset of fev-bench~\cite{shchur2025fev}, using one fixed augmentation realization per suite. The comparison roster contains seven foundation models, Chronos-2~\cite{ansari2025chronos2}, Timer-3.5~\cite{thuml2026timers1}, TimesFM 2.5~\cite{google2025timesfm25}, TimesFM 3.0~\cite{timesfm},  TiREx-2~\cite{nxai2026tirex2}, Moirai 2~\cite{salesforce2025moirai2}, and Toto 2.0~\cite{khwaja2026toto2}, plus four simple statistical baselines (Drift, Naive, Seasonal Naive, and Linear Trend). TiREx-2 is reported on GIFT-Eval Short and fev-bench Mini-20 because its output-length limit leaves others unsupported. Each Structural Feature is controlled by a salience coefficient $\lambda$, instantiated as five ordered settings L1--L5 (Appendix~\ref{app:capability-mechanisms}); we report Reference MASE and Probe MASE for endpoint accuracy, Forecast-Change NRMSE for response fidelity, and Auxiliary-Input Gap for input utility, with model support and aggregation details in Appendices~\ref{app:protocol} and~\ref{app:capability-results}.

\subsection{Accuracy and Response Fidelity}
\label{subsec:main-results}
\begin{figure}[t]
  \centering
  \IfFileExists{pics/fig_cafe_capability_radar.pdf}{%
    \includegraphics[clip,width=\textwidth]{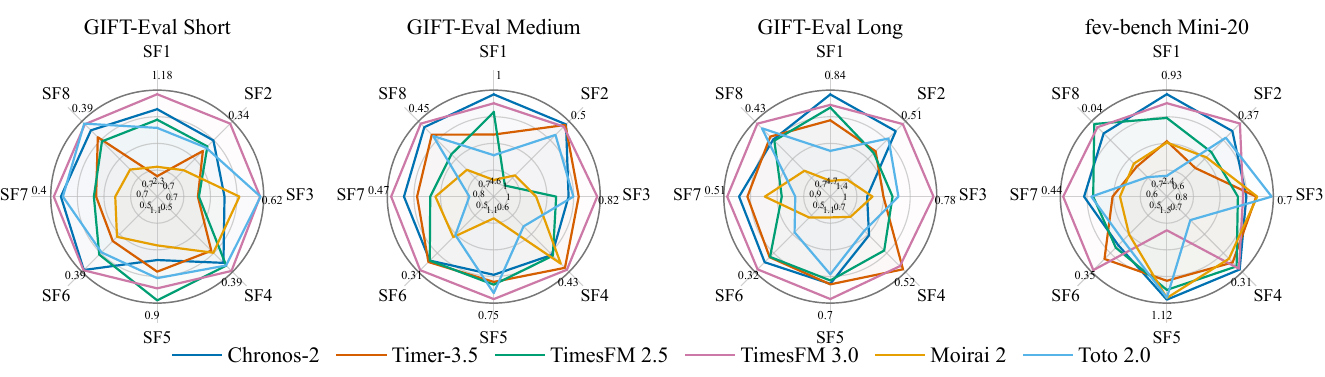}%
  }{\fbox{\parbox[c][0.31\textwidth][c]{0.92\textwidth}{\centering Feature-level response profiles}}}
  \caption{Feature-level Forecast-Change NRMSE values (lower is better) across four suites. Each SF uses an independent range over the six shared models, with larger values toward the center and smaller values on extended outer spokes; endpoint labels show both bounds.}
  \label{fig:capability-radar}
  \vspace{-8pt}
\end{figure}

CaFE reports, for each model and Structural Feature, endpoint accuracy (Probe MASE on the Refreshed Benchmark) and response fidelity (Forecast-Change NRMSE); Figures~\ref{fig:main-capability-mase-heatmaps} and~\ref{fig:main-capability-effect-heatmaps} show the two views across the four suites, and Figure~\ref{fig:capability-radar} condenses the fidelity view into a per-feature profile. Accuracy ranks models overall, while response fidelity reveals where each model's response is strong or weak: a model can be accurate overall yet respond poorly to a structural change, such as a cross-series or covariate signal. The leaders differ by what is measured, TimesFM 3.0 winning 5/8, 7/8, 5/8, and 7/8 feature tracks by Probe MASE but only 4/8, 6/8, 6/8, and 3/8 by Forecast-Change NRMSE, while Linear Trend has the lowest SF1 NRMSE despite weak endpoint accuracy.

At the aggregate level the two rankings are largely consistent: the Spearman correlation between Probe-MASE rank and NRMSE rank is $0.029$, $0.800$, $0.900$, and $0.771$ on the four suites, and the refreshed set does not distort accuracy rankings (Reference MASE and Probe MASE are rank-correlated at $0.943$, $0.700$, $0.900$, and $0.657$; Appendix~\ref{app:capability-results}). 
The complementarity therefore lies in per-feature granularity rather than a reordering of the leaderboard: response fidelity flags the specific structural conditions under which a model's endpoint accuracy does not transfer. The Linear Trend baseline exposes a further limit of the response metric: it attains the lowest SF1 NRMSE because a linear model trivially reproduces an injected linear trend, so the metric is most informative on the relational features (SF6--SF8), where matching the induced change requires modeling cross-series or covariate structure rather than reproducing a functional form.

\begin{figure}[t]
  \centering
  \IfFileExists{pics/fig_cafe_main_capability_mase_heatmaps.pdf}{%
    \includegraphics[clip,width=0.97\textwidth]{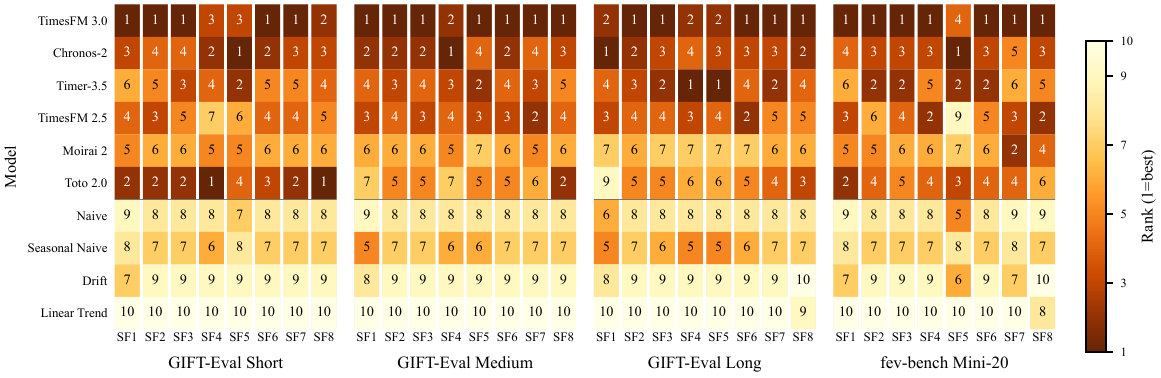}%
  }{\fbox{\parbox[c][0.29\textwidth][c]{0.92\textwidth}{\centering structural-feature Probe MASE heatmaps}}}
  \caption{Probe MASE rank by suite, model, and Structural Feature for the four suites in one horizontal strip. Within each feature, rank 1 is the most accurate model (lowest Probe MASE), and the underlying values are reported in Appendix~\ref{app:capability-results}.}
  \label{fig:main-capability-mase-heatmaps}
  \vspace{-4pt}
\end{figure}

\begin{figure}[t]
  \centering
  \IfFileExists{pics/fig_cafe_main_capability_effect_heatmaps.pdf}{%
    \includegraphics[clip,width=0.97\textwidth]{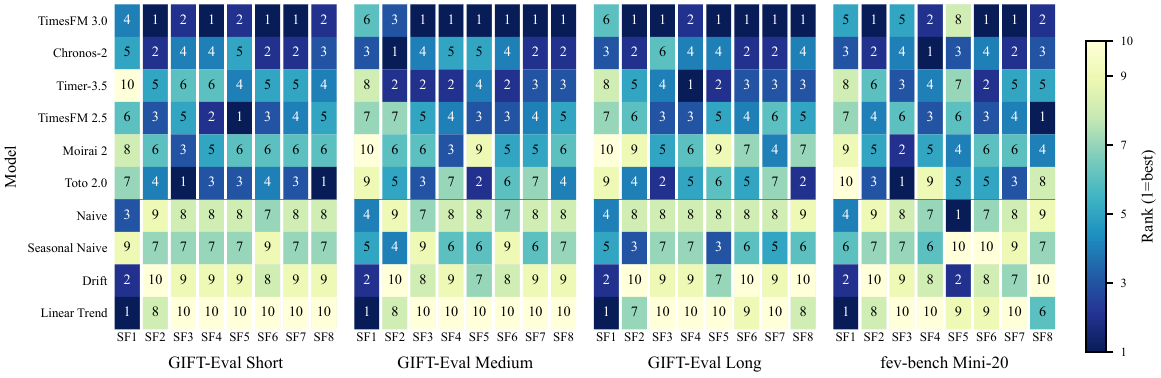}%
  }{\fbox{\parbox[c][0.29\textwidth][c]{0.92\textwidth}{\centering structural-feature Forecast-Change NRMSE heatmaps}}}
  \caption{Forecast-Change NRMSE rank by suite, model, and Structural Feature for the four suites in one horizontal strip. Within each feature, rank 1 is the most faithful response to the induced change; NRMSE~=~1 is the zero-response reference, and the underlying values are reported in Appendix~\ref{app:capability-results}.}
  \label{fig:main-capability-effect-heatmaps}
  \vspace{-10pt}
\end{figure}

\subsection{Selecting Models by Intrinsic Structure}
\label{subsec:intrinsic-selection}

\begin{wrapfigure}{R}{0.5\textwidth}
  \vspace{-12pt}
  \centering
\IfFileExists{pics/fig_cafe_intrinsic_selection.pdf}{%
    \includegraphics[width=\linewidth]{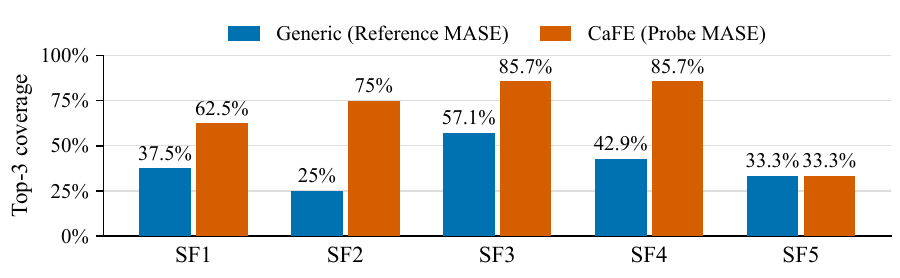}%
  }{\fbox{\parbox[c][0.24\textwidth][c]{\linewidth}{\centering Top-3 coverage bar chart}}}
  \caption{Top-3 coverage of the Reference-MASE-best model on genuine strong-structure tasks from held-out source families (higher is better). SF1--SF5 use 8, 4, 7, 7, and 3 eligible families, respectively.}
  \label{fig:intrinsic-selection}
\end{wrapfigure}

An aggregate leaderboard selects a single model for every scenario, yet the best model can depend on the target series' structural condition. For each intrinsic Structural Feature (SF1--SF5), we gather genuine tasks with strong corresponding structure, hold out one source family at a time, and compare the three models ranked highest by aggregate Reference MASE with those ranked highest by feature-specific Probe MASE. Figure~\ref{fig:intrinsic-selection} shows that CaFE raises Top-3 coverage of the Reference-MASE-best model from 41.4\% (12/29) to 72.4\% (21/29). The gain concentrates on SF1 ($37.5\% \to 62.5\%$), SF3 ($57.1\% \to 85.7\%$), and SF4 ($42.9\% \to 85.7\%$), while SF2 (multi-seasonality) and SF5 (intermittency) stay flat, consistent with their few eligible families (4 and 3) and weaker feature-local signal.

\subsection{Validation of Relational Features}
\label{subsec:relational-validation}

\begin{wrapfigure}{r}{0.54\textwidth}
  \vspace{-8pt}
  \centering
  \IfFileExists{pics/fig_cafe_model_nrmse_gap.pdf}{%
    \includegraphics[clip,width=\linewidth]{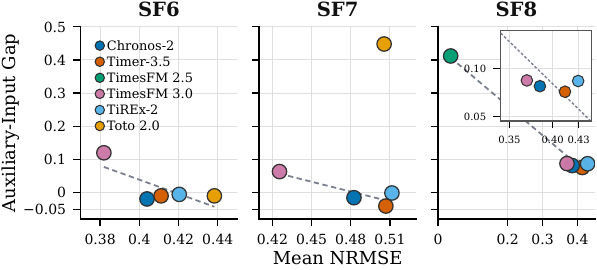}%
  }{\fbox{\parbox[c][0.18\textwidth][c]{\linewidth}{\centering model-level NRMSE--Auxiliary-Input Gap relationship}}}
  \caption{Model-level response fidelity and input utility for SF6--SF8. Each point is an equal-weight model mean across datasets; lower NRMSE and positive Gap are preferred.}
  \label{fig:model-nrmse-gap}
  \vspace{-8pt}
\end{wrapfigure}

The relational features SF6--SF8 test whether a model exploits auxiliary information: a shared latent factor across targets (SF6), a driver series (SF7), or a dynamic covariate (SF8). To validate these features, we remove the auxiliary input and measure how much the assessed target's error changes, where a positive $\Delta\mathrm{MASE}=\mathrm{MASE}_{\mathrm{ablated}}-\mathrm{MASE}_{\mathrm{full}}$ is the gain from the input. Figure~\ref{fig:model-nrmse-gap} shows this gain against response fidelity: TimesFM 3.0 has the largest SF6 gap, Toto 2.0 the largest SF7 gap, and TimesFM 2.5 the largest SF8 gap, while several models lie near or below zero, marking models for which the auxiliary input acts as a distractor. The two axes are deliberately different: the Gap measures whether a model uses an input in the original task, whereas NRMSE measures whether a forecast responds to an injected change, so their lack of a clean monotone relationship is expected rather than a failure of either metric. The gains are small and model-specific, and the strongest input gain is not consistently held by the model with the lowest NRMSE, so the value of these inputs cannot be read from a single aggregate score. As a calibration, the simple baselines that cannot exploit cross-series or covariate structure sit near the zero-response reference on SF6--SF8, whereas the foundation models that use these inputs score lower; this confirms that the response metric separates models that track the induced change from those that ignore it.


Figure~\ref{fig:model-nrmse-gap} makes the two roles visible in one plot: the horizontal axis is response fidelity and the vertical axis is input utility. In the displayed means, TimesFM 3.0 has the largest SF6 gap, Toto 2.0 the largest SF7 gap, and TimesFM 2.5 the largest SF8 gap, while several other models lie near zero or below zero. The model with the strongest input gain is therefore not consistently the model with the lowest NRMSE, and the three SFs should be read on their own supports rather than as one difficulty scale. The task counts, descriptive intervals and details are reported in Appendix~\ref{app:capability-results}.


\subsection{Faithfulness of the Augmentation}
\label{subsec:faithfulness}

\begin{wrapfigure}{R}{0.45\textwidth}
  \vspace{-4pt}
  \centering
  \IfFileExists{pics/rank_agreement.pdf}{%
    \includegraphics[clip,width=\linewidth]{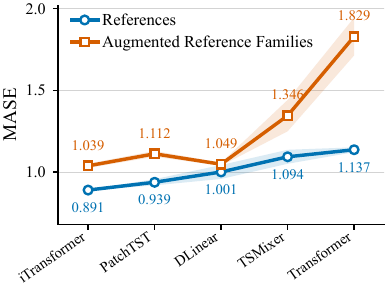}%
  }{\fbox{\parbox[c][0.24\textwidth][c]{\linewidth}{\centering paired original--treatment rank agreement}}}
  \caption{MASE on the References and on the Augmented Reference Families for five small models, ordered by Reference MASE. Points are three-seed means with $\pm1$ sample SD; near-identical pairs within a model indicate that augmentation preserves that model's accuracy ordering.}
  \label{fig:rank-agreement}
  \vspace{-8pt}
\end{wrapfigure}

A useful augmentation extends a benchmark without distorting model ordering; otherwise the structural signals CaFE adds would be noise rather than faithful evaluation. We verify this by ranking five small models on the References and the augmented Reference Families. Figure~\ref{fig:rank-agreement} shows the paired Reference and Augmented-Reference-Family MASE across 2,436 common source pairs and 21,793 valid augmented samples: agreement is Spearman $\rho=0.90$ and Kendall $\tau_b=0.80$, with nine of ten pairwise orderings preserved, though heterogeneous across groups ($\rho=1.00$ multivariate, $0.30$ univariate, $-0.10$ Trend; Appendix~\ref{app:small-model-protocol}). We therefore treat this as conditional evidence for paired rank stability on this suite, not universal rank preservation. The single exception is Trend ($\rho=-0.10$), the same phenomenon seen in Section~\ref{subsec:main-results}: a linear intervention is nearly trivial to reproduce, so the Trend feature is the least discriminative and the least order-preserving; faithfulness is strongest where the paired change carries genuine relational information.


\subsection{Stability under Benchmark Refresh}
\label{subsec:stability}

Because CaFE is refreshable, its conclusions should survive regeneration. Regenerating the GIFT-Eval Short benchmark five times, changing only the augmentation seed, leaves the overall ranking unchanged, with TimesFM 3.0 first on both metrics (Table~\ref{tab:five-seed-stability}); the scores vary by 0.26\%--0.51\% (Probe MASE) and 0.59\%--1.22\% (Forecast-Change NRMSE). Feature-level winners are less stable, the lowest-NRMSE model changing across seeds on SF3, SF6, and SF8. Overall ordering is thus robust to regeneration, while feature-level winners reveal where a comparison depends on the generated conditions. This feature-level sensitivity means the per-feature winners reported in Section~\ref{subsec:main-results} are themselves seed-dependent; the robust finding is the stable overall ordering, while the specific per-feature leaders should be read as indicative rather than fixed.


\subsection{Tradeoff between Two Metrics}
\label{subsec:tradeoff}

\begin{wrapfigure}{R}{0.4\textwidth}
  \vspace{-12pt}
  \centering
  \caption{Five-seed GIFT-Eval Short stability on the generated cohorts. Values are mean $\pm$ seed SD of Probe MASE and Forecast-Change NRMSE; ordering is unchanged across five seeds.}
  \label{tab:five-seed-stability}
  \scriptsize
  \setlength{\tabcolsep}{2pt}
  \renewcommand{\arraystretch}{1.1}
  \resizebox{\linewidth}{!}{%
    \begin{tabular}{@{}lcc@{}}
      \toprule
      Model & Probe MASE & \shortstack{Forecast-Change\\NRMSE} \\
      \midrule
      TimesFM 3.0  & 1.21 $\pm$ 0.003 & 0.580 $\pm$ 0.005 \\
      TiREx-2      & 1.20 $\pm$ 0.003 & 0.637 $\pm$ 0.007 \\
      Chronos-2    & 1.22 $\pm$ 0.004 & 0.646 $\pm$ 0.007 \\
      Toto 2.0     & 1.24 $\pm$ 0.004 & 0.671 $\pm$ 0.004 \\
      TimesFM 2.5  & 1.25 $\pm$ 0.004 & 0.684 $\pm$ 0.008 \\
      Timer-3.5    & 1.28 $\pm$ 0.005 & 0.788 $\pm$ 0.009 \\
      Moirai 2     & 1.30 $\pm$ 0.005 & 0.805 $\pm$ 0.010 \\
      \bottomrule
    \end{tabular}
  }
  \vspace{-8pt}
\end{wrapfigure}

\begin{wrapfigure}{R}{0.595\textwidth}
  \vspace{-6pt}
  \centering
  \IfFileExists{pics/fig_cafe_finetuning_seed_a.pdf}{%
    \includegraphics[clip,width=\linewidth]{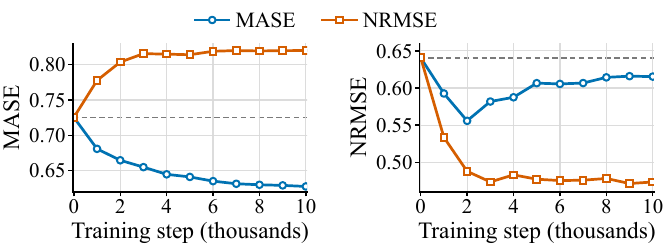}%
  }{\fbox{\parbox[c][0.17\textwidth][c]{\linewidth}{\centering Fine-tuning trade-off}}}
  \caption{Chronos-2 fine-tuned on 11,269 \textsc{CaFE}-generated samples under two objectives. The two panels track MASE and Forecast-Change NRMSE over 0--10k steps; official quantile-loss training improves MASE, while the Forecast-Change-NRMSE objective improves response fidelity at the cost of endpoint accuracy.}
  \label{fig:single-seed-finetuning}
  \vspace{-6pt}
\end{wrapfigure}

The two metrics also move in opposite directions when the training objective changes. Fine-tuning Chronos-2 on the same 11,269 CaFE-generated samples, official quantile-loss training lowers MASE by 13.46\% but improves NRMSE by only 3.96\%, whereas the Forecast-Change-NRMSE objective lowers NRMSE by 26.12\% while raising MASE by 13.04\% (Figure~\ref{fig:single-seed-finetuning}). The training objective therefore trades endpoint accuracy against response fidelity, so a practitioner can steer the model toward the capability their scenario needs. This is a single training realization. Optimizing Forecast-Change NRMSE teaches the model to track the injected changes, which trades away the endpoint fit on the refreshed distribution; the two objectives therefore move a single accuracy--fidelity operating point rather than demonstrating that one objective dominates the other.

\section{Discussion}
\label{sec:discussion}
The novelty of CaFE is that a single, controlled, salience-varying perturbation of an unchanged Reference yields two measurements at once: endpoint accuracy and response fidelity. This paired Reference--sample design is what separates CaFE from fixed leaderboards, which report only endpoint accuracy, and from synthetic benchmarks, which replace the source task. Because every intervention stays anchored to a real instance, one refreshed benchmark produces a per-feature capability profile alongside the leaderboard, so the contribution is a capability-oriented axis that is grounded in, rather than divorced from, the original evaluation.

Across GIFT-Eval and fev-bench Mini-20, the two metrics reveal condition-specific strengths and weaknesses that a single aggregate score hides, the augmentation preserves model ordering, and the rankings are stable under regeneration. Most importantly, on genuine data filtered by intrinsic structure, models selected by CaFE's per-feature profiles outperform the aggregate leaderboard pick, so structural response can guide model selection where a single score cannot.
Several limits remain. Stability rests on five seeds on one suite, fine-tuning on a single realization, and the relational validation has few significant cells; selection is shown only for intrinsic structures. Future work includes extending to relational and covariate conditions, more sources, and multi-seed recipes, along with controlled calibration of the response metric.

\section{Conclusion}
\label{sec:conclusion}
CaFE extends static time series foundation model benchmarks with refreshable structural diversity and reports endpoint accuracy, response fidelity, and auxiliary-input utility as separate views. On genuine data filtered by intrinsic structure, models selected by CaFE outperform the aggregate leaderboard pick, so model selection can be made conditional on the structural response of interest. The augmentation preserves model ordering, the rankings are stable under regeneration, and endpoint accuracy and response fidelity trade off under a changed training objective.

\clearpage
\section*{AI Use Disclosure}
Generative AI tools were used to assist with limited aspects of this work, including code implementation support, experiment analysis, and manuscript refinement. These tools were used for tasks such as language editing, text organization, formatting conversion, compilation and layout checks, and consistency verification of terminology and figures. We did not use generative AI tools to generate datasets, experimental results, or mathematical claims. All AI-assisted outputs were reviewed and validated by the authors, who take full responsibility for the final manuscript, including all text, claims, analyses, figures, code, and research artifacts.
\section*{Reproducibility Statement}
The anonymous artifact, including code and compact numerical inputs for the reported analyses, is available at \url{https://anonymous.4open.science/r/CaFE-ICLR2027-5210}. Appendix~\ref{app:protocol} specifies data construction, scoring, and aggregation; Appendix~\ref{app:evaluation-coverage} records the evaluation panels, model support, and evidence scope; and Appendix~\ref{app:reproduction-guide} documents the computing environments and procedures for reconstructing the tables and figures and rerunning the evaluation. The reported results use the declared panels and one training realization.

\bibliographystyle{iclr2027_conference}
\bibliography{refs}
\clearpage
\appendix
\section{Profiling, Augmentation, and Structural Feature Constructions}
\label{app:capability-mechanisms}

The contracts below define task-qualified References and the eight Structural Feature constructions. The admitted GIFT-Eval tasks form a narrower population than the complete official benchmark; Appendix~\ref{app:capability-results} reports the support of each analysis.

\subsection{Stage 1: Profiling}
\label{app:references}

\subsubsection{Task contract for each benchmark}

Let $t$ denote one benchmark task.  Its adapter contract identifies the source data and fixes the forecast targets and their order, forecast-origin enumeration and horizon, the input history in the benchmark's native representation, auxiliary inputs allowed by the task and their visibility roles, observation policy, and validity conditions.  A univariate setting has one prediction target, whereas a multivariate setting preserves multiple ordered targets as one forecasting object.  Either target setting is covariate-informed when the task additionally provides non-target covariates allowed by the task.  The adapter maps the native representation to CaFE's common interface while preserving these semantics; any model-required channel decomposition is recorded as an inference adaptation rather than a change to the task's target setting.

For the $i$th valid forecast origin of task $t$, the adapter returns a \emph{Reference}, an adapter-qualified forecast instance from the original benchmark task in \textsc{CaFE}'s common representation, written as
\begin{equation}
 R_i=\left(H_i,Y_i^{+},Z_i,M_i,o_i,\iota_i\right).
 \label{eq:app-reference}
\end{equation}
Here, $H_i$ is the target history, $Y_i^{+}$ is the forecast truth, $Z_i$ contains the auxiliary inputs permitted by $t$, $M_i$ records the applicable observation masks, including the future-target observation mask that identifies scoreable cells, $o_i$ is the forecast origin, and $\iota_i$ identifies the benchmark task and native window used for the forecast. The task definition provides the information needed to reconstruct the corresponding Reference.  The References of task $t$ form
\begin{equation}
 \mathcal A_t=\{R_i:R_i\text{ satisfies the adapter contract for }t\}.
 \label{eq:app-reference-set}
\end{equation}

\paragraph{Reference materialization and information boundary.} The task adapter owns forecast-origin enumeration and future-label eligibility; it may require a fully observed horizon or retain a future-target observation mask, with the resulting scoreability state carried by the Reference. History repair, when required, follows a deterministic adapter-resolved rule derived from history, and the corresponding scale policy declares whether repaired values or originally observed pairs enter history-based statistics. The source history, task contract, and recorded augmentation seed determine Structural Feature selection and every structural, control-setting, and random choice; future target values enter the workflow as scoreable truth after these choices. The future-target observation mask may constrain where a constructed effect must remain visible and determines which future cells are scoreable. Horizon visibility follows the declaration of task $t$: known-future dynamic covariates extend across the horizon, and past-only fields end at the forecast origin. Profiling records the adapter, task, source, and native-window identities together with the bound task contract for each $R_i$, supporting later contract checks and source reconstruction under the verification policy selected for the evaluation round.

\subsection{Stage 2: Augmentation}
\label{app:capability-units}

\subsubsection{Contract for an ordered set of samples}

For Structural Feature $s\in\{1,\ldots,8\}$, Augmentation first applies an SF-specific source filter $F_s^{\mathrm{pre}}$ to References from the original benchmark.  Its inputs are source history, Structural Feature evidence, and the task contract.  It retains $R_i$ only when the selected SF has sufficient source support, every declared setting admits a feasible construction that can satisfy the Structural Feature Expectation, and the inputs and construction domain are physically valid.  The retained source Reference is written $R_i'$ to mark the selection step; the prime changes no task field or provenance.  For each selected Reference $R_i'$ and Structural Feature $s$, Augmentation selects a shared structural configuration $\xi_{is}$.  It fixes the affected targets and feature-specific temporal form or predictive relation across the ordered index set $\mathcal L_s$.  The evaluation round declares this set; the experimental setting counts and parameter ranges appear in Appendix~\ref{app:capability-results} and Appendix~\ref{app:capability-mechanisms}.  The index $\ell\in\mathcal L_s$ marks a position within $s$, and $\lambda_{is\ell}$ is the setting of the feature-specific Control Parameter.  The corresponding generated sample and sample set are

\begin{equation}
 \begin{aligned}
 V_{is\ell}
 &=T_s\!\left(R_i';\xi_{is},\lambda_{is\ell}\right),\\
 \widetilde{\mathcal V}_{is}
 &=\{V_{is\ell}:\ell\in\mathcal L_s\},
 \end{aligned}
 \label{eq:app-generated-sample-set}
\end{equation}
where $T_s$ changes the feature-declared fields while preserving the task, source identifier, origin, horizon, target order, visibility roles, and shared configuration $\xi_{is}$. The Control Parameter may represent amplitude, change location, period count, or event spacing, so its setting order and scale are local to each Structural Feature. Applying a setting may determine compatible details such as the phase used by SF5; all settings retain $R_i'$ as their common baseline.

The protocol permits source selection to resample existing References or structural regions according to a declared Structural Feature Distribution.  Resampling changes the structural conditions supplied to Augmentation and records the selected view $R_i'$.  The unchanged Reference population remains the estimand for original-task results, and each run records how its selection policy is instantiated.

\paragraph{Field-level transformation contract.}
The feature-specific transformation can be written componentwise as
\begin{equation}
 \begin{aligned}
 V_{is\ell}=\bigl(&H_i+\Delta H_{is\ell},\;Y_i^{+}+\Delta Y_{is\ell},\\
                  &Z_i+\Delta Z_{is\ell},\;M_i,\;o_i,\;\iota_i\bigr).
 \end{aligned}
 \label{eq:app-field-level-transformation}
\end{equation}
The deltas identify the fields changed by Structural Feature $s$; zero deltas retain the corresponding selected-Reference fields.  SF1--SF7 act on target histories and their corresponding generated future targets, so $\Delta Z_{is\ell}=0$.  SF8 adds a dynamic-covariate impulse allowed by the task and its corresponding target response, with its future covariate component restricted by the task's visibility declaration.  Thus, every transformation changes only the feature-declared fields of $R_i'$; Evaluation measures the resulting forecast changes.

\paragraph{Structure selection and setting feasibility.} Each mechanism determines applicability and constructs a feasible shared structure. For amplitude-controlled features, the generator normalizes the selected component by its source distance and selects settings only from the feasible interval; SF2, SF4, and SF5 vary period count, transition location, and event spacing. The selected structure is fixed across the ordered settings.

\paragraph{Source-distance evidence.}
Qualification uses a common distance to keep each generated change visible and bounded relative to its source.  Let $\mathcal J_{is}$ be the affected target set, $C$ a protocol-declared history view, and $\sigma_{id}>0$ the source-history standard deviation for target $d$.  A nonfinite or degenerate standard deviation falls back first to the standard deviation of first differences and then to the declared positive value $1$.  This quantity normalizes construction distance, while the MASE scale $\sigma^{\mathrm{MASE}}_{id}$ normalizes forecast error.  Stage~2 records
\begin{equation}
 d_{is\ell C}
 =\frac{1}{|\mathcal J_{is}|}
 \sum_{d\in\mathcal J_{is}}
 \sqrt{
   \frac{1}{|C|}
   \sum_{u\in C}
   \left(
     \frac{H^{V}_{is\ell,u,d}-H^{R}_{i,u,d}}{\sigma_{id}}
   \right)^2
 }.
 \label{eq:app-source-distance}
\end{equation}
This distance uses the selected Reference $R_i'$ as the comparator for every declared setting.  For every required model-visible context, the macro value must lie in $[0.10,2.0]$; each affected channel has an upper bound of $3.0$.  Structural variables such as period count or event spacing serve as feature-specific Control Parameters and express structural progression.

\subsection{Structural Feature Constructions}
\label{app:sf-constructions}

\begin{table}[t]
  \centering
\caption{Structural features, controlled changes, control parameters, and input requirements. Full constructions appear in this appendix.}
  \label{tab:structural-features}
  \scriptsize
  \setlength{\tabcolsep}{4pt}
  \begin{tabularx}{\textwidth}{@{}p{0.23\textwidth}Xp{0.23\textwidth}p{0.14\textwidth}@{}}
    \toprule
    Structural feature & Controlled change & Control parameter & Applicability \\
    \midrule
    SF1 Trend & Persistent directional continuation & Normalized trend magnitude & Uni./multi. \\
    SF2 Multi-Seasonality & Concurrent stable periods & Number of active periods & Uni./multi. \\
    SF3 Time-Varying Seasonality & Evolving amplitude or phase & Normalized evolution magnitude & Uni./multi. \\
    SF4 Regime Switching & Bounded transition to a persistent state & Transition location & Uni./multi. \\
    SF5 Intermittency & Sparse recurring events & Mean event gap & Uni./multi. \\
    SF6 Common Factor & Shared trajectory across targets & Normalized shared-profile magnitude & Multi., $\geq3$ targets \\
    SF7 Cross-Series Dependence & Directed driver--responder relation & Normalized transfer magnitude & Multi., $\geq2$ targets \\
    SF8 Covariate Response & Covariate impulse with matched target response & Joint normalized response magnitude & Native dynamic covariate \\
    \bottomrule
  \end{tabularx}
\end{table}

Table~\ref{tab:structural-features} summarizes the eight constructions. The reported GIFT-Eval and fev-bench Mini-20 experiments use $\mathcal L_s=\{1,\ldots,5\}$, displayed as L1--L5. The formulas specify the changed fields, shared structural configuration, and feature-specific control settings; the numerical ranges below belong to this experimental configuration.

\subsection{Shared Notation and Control Contract}

For Reference $R_i$, let $L_i$ and $H_i^{\mathrm{hor}}$ denote its history length and forecast horizon.  We use $u\in\{0,\ldots,L_i+H_i^{\mathrm{hor}}-1\}$ on the concatenated target path
\begin{equation}
 X_{i,d}(u)=
 \begin{cases}
 H_i[u,d], & 0\le u<L_i,\\
 Y_i^{+}[u-L_i,d], & L_i\le u<L_i+H_i^{\mathrm{hor}},
 \end{cases}
 \label{eq:app-complete-target-path}
\end{equation}
where $d$ indexes prediction targets.  Let $U_{i,q}(u)$ denote task-permitted dynamic covariate $q$ on its declared visible range.  A generated sample changes these fields by
\begin{equation}
 X^{V}_{is\ell}=X_i+\Delta X_{is\ell},
 \qquad
 U^{V}_{is\ell}=U_i+\Delta U_{is\ell}.
 \label{eq:app-dimension-field-update}
\end{equation}
The future values in Equation~\eqref{eq:app-complete-target-path} provide scoreable truth after construction. Structural configuration $\xi_{is}$ is selected from source history and task information before those values enter the workflow.

For the amplitude-controlled Structural Features SF1, SF3, SF6, SF7, and SF8, let $G^X_{is}(u,d)$ and $G^U_{is}(u,q)$ be the raw target and covariate components fixed by $\xi_{is}$.  With affected-target set $\mathcal J_{is}$ and source-history standard-deviation scale $\sigma_{id}$, define the full-history macro normalized root-mean-square
\begin{equation}
 \mathcal D_i(G^X_{is};\mathcal J_{is})
 =\frac{1}{|\mathcal J_{is}|}
 \sum_{d\in\mathcal J_{is}}
 \sqrt{\frac{1}{L_i}\sum_{u=0}^{L_i-1}
 \left(\frac{G^X_{is}(u,d)}{\sigma_{id}}\right)^2}.
 \label{eq:app-component-distance}
\end{equation}
Writing $q_{is}=\mathcal D_i(G^X_{is};\mathcal J_{is})$, Salience Tuning at control setting $\ell\in\mathcal L_s$ gives
\begin{equation}
 \left(\Delta X_{is\ell},\Delta U_{is\ell}\right)
 =\frac{\lambda_{is\ell}}{q_{is}}
 \left(G^X_{is},G^U_{is}\right).
 \label{eq:app-amplitude-level-map}
\end{equation}
In this experimental instantiation, the five nominal intervals for $\lambda_{is\ell}$ are $[0.10,0.15]$, $[0.17,0.22]$, $[0.25,0.32]$, $[0.36,0.46]$, and $[0.50,0.65]$, intersected with the feasible interval of the selected structure.  SF2, SF4, and SF5 instead use period count, transition location, and event gap as their feature-specific Control Parameters.  Each SF therefore interprets L1--L5 through its own Control Parameter and Control Settings, and Evaluation measures the forecast changes under those settings.

\begin{figure}[t]
  \centering
  \includegraphics[clip,width=0.99\textwidth]{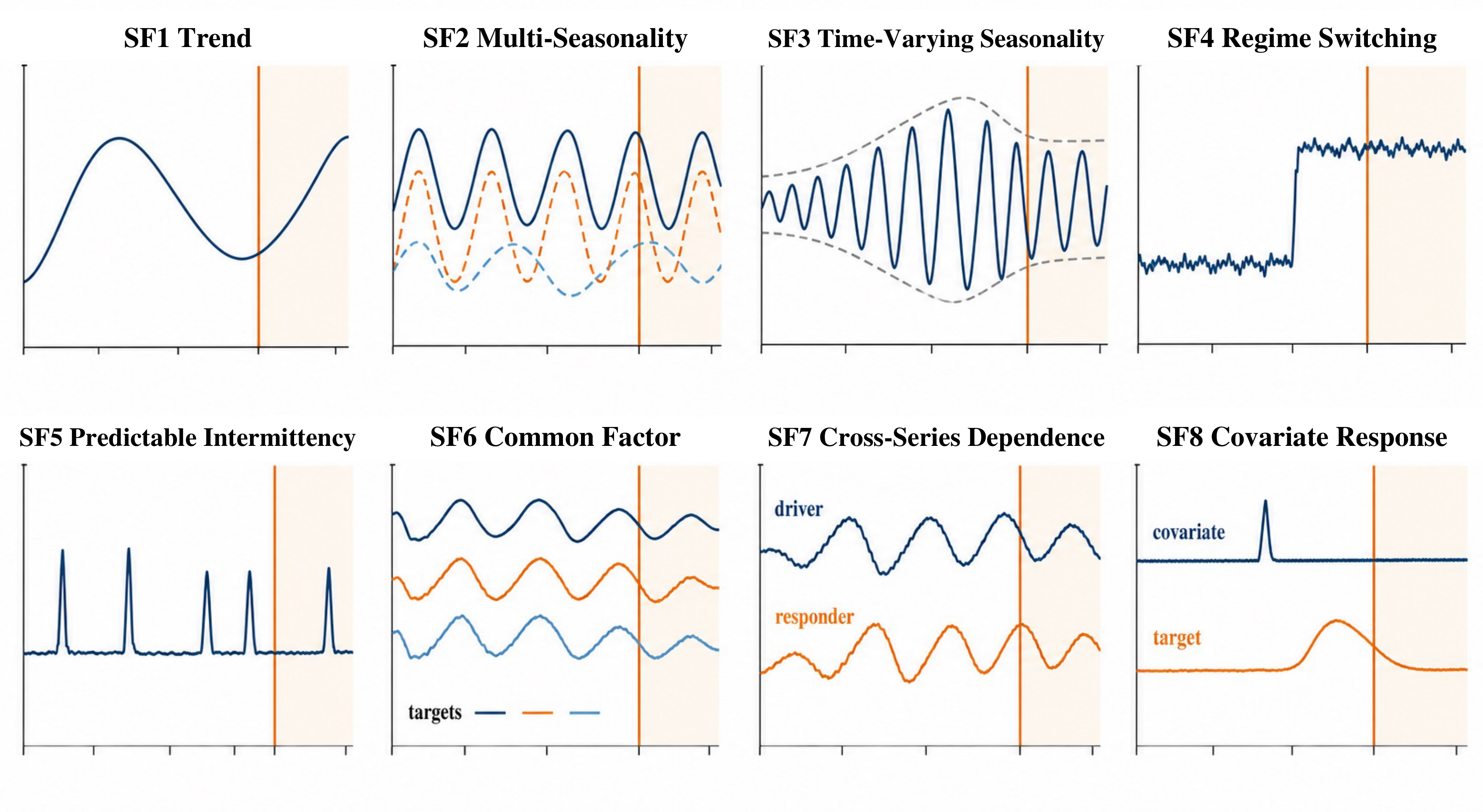}
  \caption{Illustrative interventions for SF1--SF8. Each panel joins the modified history to its analytic continuation over the forecast horizon and preserves the task\textquotesingle s target and auxiliary-input roles.}
  \label{fig:capability-mechanism-examples}
\end{figure}

Figure~\ref{fig:capability-mechanism-examples} illustrates the constructed components defined below.

\subsection{SF1 Trend}
\label{app:capability-trend}

SF1 generates a trend by selecting a nonempty target subset $\mathcal J_{i1}$ with history-stable trend signs $s_{id}^{\mathrm{tr}}\in\{-1,1\}$.  For each selected target, the raw component is
\begin{equation}
 G^X_{i1}(u,d)
 =\mathbf 1\{d\in\mathcal J_{i1}\}\,
 \sigma_{id}s_{id}^{\mathrm{tr}}p_{i1}(u),
 \qquad
 \Delta X_{i1\ell}=\frac{\lambda_{i1\ell}}{q_{i1}}G^X_{i1},
 \label{eq:app-sf1-trend}
\end{equation}
where $p_{i1}$ is one structure shared across all settings:
\begin{equation}
 \begin{aligned}
 p_{\mathrm{lin}}(u)&=\frac{u}{L_i-1},\\
 p_{\mathrm{del}}(u)&=\frac{(u-u_0)_+}{L_i-1},\\
 p_{\mathrm{curv}}(u)&=\left(\frac{u}{L_i-1}\right)^r,\\
 p_{\mathrm{pw}}(u)&=\frac{1}{L_i-1}\sum_{v=0}^{u-1}
 b_{\left\lfloor(v+\phi)/K\right\rfloor\bmod B},\\
 p_{i1}&\in\{p_{\mathrm{lin}},p_{\mathrm{del}},p_{\mathrm{curv}},p_{\mathrm{pw}}\}.
 \end{aligned}
 \label{eq:app-sf1-profile-family}
\end{equation}
The four alternatives are whole-history linear, delayed linear, slow-curvature, and recurring piecewise-linear continuation.  The onset $u_0$, curvature $r$, knot interval $K$, phase $\phi$, slope cycle $(b_0,\ldots,b_{B-1})$, affected targets, and signs belong to $\xi_{i1}$; only normalized magnitude $\lambda_{i1\ell}$ changes across settings.  Construction requires a stable history-derived direction and continues the same profile law through the horizon.

\subsection{SF2 Multi-Seasonality}
\label{app:capability-multi-seasonal}

For each target, SF2 constructs multi-seasonality by selecting a harmonic anchor from up to three qualified history-spectrum candidates, with a protocol-generated anchor when none qualifies. For target $d$, let the selected anchor and five additional independent periods be $P_{id0},\ldots,P_{id5}$.  Component $r$ is either a sinusoid or a two-harmonic waveform,
\begin{equation}
 \begin{aligned}
 h_{idr}(u)
 &=\sum_{m=1}^{M_{idr}}a_{idrm}
 \sin\!\left(\frac{2\pi m u}{P_{idr}}\right)\\
 &\quad+\sum_{m=1}^{M_{idr}}b_{idrm}
 \cos\!\left(\frac{2\pi m u}{P_{idr}}\right),\\
 M_{idr}&\in\{1,2\}.
 \end{aligned}
 \label{eq:app-sf2-component}
\end{equation}
Each component is history-normalized to the target scale before aggregation.  At Control Setting $\ell\in\{1,\ldots,5\}$, the generated sample contains the anchor and the first $\ell$ additional periods:
\begin{equation}
 S_{i2\ell}(u,d)=\sum_{r=0}^{\ell}h_{idr}(u),
 \qquad
 \Delta X_{i2\ell}(u,d)
 =\frac{\rho_{i2}}{\mathcal D_i(S_{i2\ell};\mathcal J_{i2})}
 S_{i2\ell}(u,d),
 \label{eq:app-sf2-multi-seasonality}
\end{equation}
where $\mathcal J_{i2}$ contains all targets and the shared full-history distance $\rho_{i2}\in[0.22,0.28]$ is fixed across the five settings.  Hence the Control Parameter is the number of additional independent periods, producing two through six periods while holding aggregate history amplitude fixed.  Periods, phases, waveform families, and their order are selected from history-qualified or protocol-generated candidates without future target values.

\subsection{SF3 Time-Varying Seasonality}
\label{app:capability-time-varying-seasonality}

SF3 constructs time-varying seasonality.  For an eligible target $d$, the selected carrier, its quadrature component, and a slower envelope are
\begin{equation}
 \begin{aligned}
 c_{id}(u)&=a_{id}^{c}\sin(\omega_{id}^{c}u)+b_{id}^{c}\cos(\omega_{id}^{c}u),\\
 c_{id}^{\perp}(u)&=b_{id}^{c}\sin(\omega_{id}^{c}u)-a_{id}^{c}\cos(\omega_{id}^{c}u),\\
 e_{id}(u)&=a_{id}^{e}\sin(\omega_{id}^{e}u)+b_{id}^{e}\cos(\omega_{id}^{e}u),
 \qquad \omega_{id}^{e}<\omega_{id}^{c}.
 \end{aligned}
 \label{eq:app-sf3-carrier-envelope}
\end{equation}
The raw component uses the carrier for amplitude modulation and the quadrature carrier for the periodic phase-drift component:
\begin{equation}
 \begin{aligned}
 b_{id}^{\mathrm{mod}}(u)&=
 \begin{cases}
 c_{id}(u), & \text{amplitude modulation},\\
 c_{id}^{\perp}(u), & \text{periodic phase drift},
 \end{cases}\\
 G^X_{i3}(u,d)&=\mathbf 1\{d\in\mathcal J_{i3}\}\,
 e_{id}(u)b_{id}^{\mathrm{mod}}(u),\\
 \Delta X_{i3\ell}&=\frac{\lambda_{i3\ell}}{q_{i3}}G^X_{i3}.
 \end{aligned}
 \label{eq:app-sf3-time-varying-seasonality}
\end{equation}
History fits determine the carrier and slower envelope; for each eligible target, the seed selects one qualified carrier and modulation from a leading pool of at most eight, then selects a nonempty target subset.  Control Setting $\ell$ changes only the normalized magnitude.  Qualification requires the envelope to remain active on the declared fraction of observed horizon positions.

\subsection{SF4 Regime Switching}
\label{app:capability-regime-switching}

SF4 constructs regime switching.  Let $\tau_{i4\ell}$ be the setting-specific transition location, $w_{i4}$ the shared transition width, and
\begin{equation}
 \begin{aligned}
 p(u)&=\operatorname{clip}\!\left(\frac{u-\tau}{w},0,1\right),\\
 r_{\kappa}(u;\tau,w)&=
 \begin{cases}
 \mathbf 1\{u\ge\tau\}, & \kappa=\mathrm{step},\\
 p(u), & \kappa=\mathrm{ramp},\\
 3p(u)^2-2p(u)^3, & \kappa=\mathrm{sigmoid},
 \end{cases}
 \end{aligned}
 \label{eq:app-sf4-transition-profile}
\end{equation}
With affected-target signs $s_{id}^{\mathrm{rg}}$, the five treatments are
\begin{equation}
 \Delta X_{i4\ell}(u,d)
 =\alpha_{i4}\,\mathbf 1\{d\in\mathcal J_{i4}\}\,
 \sigma_{id}s_{id}^{\mathrm{rg}}
 r_{\kappa_{i4}}(u;\tau_{i4\ell},w_{i4}).
 \label{eq:app-sf4-regime-switching}
\end{equation}
  The seed fixes $\mathcal J_{i4}$, the signs, transition family $\kappa_{i4}$, and width.  The Control Parameter is the transition location, represented by the ordered settings $\tau_{i4\ell}/L_i$; larger settings place the same state change closer to the forecast origin and therefore expose less post-transition history.  The shared gain $\alpha_{i4}$ is chosen once for the five settings to meet the history-distance contract.

\subsection{SF5 Intermittency}
\label{app:capability-intermittency}

SF5 constructs intermittency.  Let $w_{i5}$ be the shared pulse width.  For offset $r$, the selected pulse kernel is
\begin{equation}
 \kappa_{i5}(r)=
 \begin{cases}
 1, & 0\le r<w_{i5},\quad \text{rectangular},\\
 1-\dfrac{|r-(w_{i5}-1)/2|}{\max(1,(w_{i5}+1)/2)},
   & 0\le r<w_{i5},\quad \text{triangular},\\
 \exp\!\left[-\dfrac{(\log 2)r}{\max(1,w_{i5}/2)}\right],
   & 0\le r<w_{i5},\quad \text{exponential},\\
 0, & \text{otherwise}.
 \end{cases}
 \label{eq:app-sf5-pulse-kernel}
\end{equation}
For Control Setting $\ell$, event centers $\mathcal E_{i5\ell}$ follow mean gap $g_{i5\ell}$ with bounded alternating jitter $\epsilon_{i5\ell}$, so consecutive gaps are $\max(w_{i5}+1,g_{i5\ell}\pm\epsilon_{i5\ell})$.  The target increment is
\begin{equation}
 \Delta X_{i5\ell}(u,d)
 =\alpha_{i5}\,\mathbf 1\{d\in\mathcal J_{i5}\}\,
 \sigma_{id}s_{id}^{\mathrm{ev}}
 \sum_{t\in\mathcal E_{i5\ell}}\kappa_{i5}(u-t).
 \label{eq:app-sf5-intermittency}
\end{equation}
The shared structure fixes the affected targets, event signs, pulse family, width, and jitter fraction.  The Control Parameter is the mean-gap fraction of the maximum legal gap; larger settings are sparser.  A deterministic compatible phase retains at least three history events and an event on an observed future position.  The common gain $\alpha_{i5}$ is fixed across settings and satisfies the history-distance and minimum future-effect gates.

\subsection{SF6 Common Factor}
\label{app:capability-common-factor}

SF6 constructs common factor.  Standardize the target history as $Z_i(u,d)=(H_i[u,d]-\mu_{id})/\sigma_{id}$ and write its singular value decomposition as $Z_i=U_i\Sigma_iV_i^{\top}$.  Candidate loading vectors are masked and renormalized members of
\begin{equation}
 \begin{aligned}
 \mathcal B_i&=\{v_1,\ v_1\pm0.55v_2,\ v_1\pm0.45v_3\},\\
 \mathcal W_i&=\{\operatorname{mask}_{\eta}(w):
 w\in\mathcal B_i,\ \eta\in\{0.10,0.20,0.35\}\}.
 \end{aligned}
 \label{eq:app-sf6-loading-pool}
\end{equation}
where $v_j$ is row $j$ of $V_i^{\top}$ and $\operatorname{mask}_{\eta}$ removes loadings below fraction $\eta$ of the largest absolute loading while retaining at least three targets.  For a selected loading $w_{id}$, the history-fitted factor $Z_iw_i$ supplies a stable harmonic carrier $f_i(u)=a_i^f\sin(\omega_i^fu)+b_i^f\cos(\omega_i^fu)$.  The controlled component is
\begin{equation}
 G^X_{i6}(u,d)=\mathbf 1\{d\in\mathcal J_{i6}\}\,
 \sigma_{id}w_{id}f_i(u),
 \qquad
 \Delta X_{i6\ell}=\frac{\lambda_{i6\ell}}{q_{i6}}G^X_{i6}.
 \label{eq:app-sf6-common-factor}
\end{equation}
After qualification and deduplication, the seed selects one of at most 12 leading loading--carrier candidates; the loading, participating targets, and carrier remain fixed across settings.  The Control Parameter is normalized common-factor magnitude.  This construction measures observable co-movement under the selected loading and carrier.

\subsection{SF7 Cross-Series Dependence}
\label{app:capability-cross-series}

SF7 constructs lagged dependence by enumerating ordered driver--responder pairs and positive lags up to 24 or one eighth of the Reference-history length, whichever is smaller. For each ordered driver $p$, responder $d$, and positive lag $\tau$, SF7 fits the history-only predictive comparison
\begin{equation}
 \begin{aligned}
 z_{i,d}(u)
 &=\beta_0+\beta_1z_{i,d}(u-1)+\beta_2z_{i,p}(u-\tau)+\varepsilon_u,\\
 G^X_{i7}(u,j)
 &=\mathbf 1\{j=d\}\,\sigma_{id}\beta_2\widetilde z_{i,p}(u-\tau),\\
 \Delta X_{i7\ell}
 &=\frac{\lambda_{i7\ell}}{q_{i7}}G^X_{i7}.
 \end{aligned}
 \label{eq:app-sf7-cross-series}
\end{equation}
Here $z_{i,d}$ is the standardized target history, and $\widetilde z_{i,p}$ equals the observed driver history before the forecast origin and its history-fitted linear continuation afterward.  Candidate edges must improve predictive fit over the responder-only model. SF7 ranks them by incremental predictive fit, scans at most 48 edges, and retains at most 12 that satisfy source-distance feasibility; the seed selects one retained edge.  The driver, responder, lag, and coefficient $\beta_2$ are fixed across settings, while $\lambda_{i7\ell}$ controls transfer magnitude.  Under this history-only construction, the resulting edge is interpreted as a predictive diagnostic for SF7.

\subsection{SF8 Covariate Response}
\label{app:capability-covariate-response}

SF8 constructs covariate response by selecting a legal dynamic covariate $q$, target $d$, event set $\mathcal E_{i8}$, covariate sign $s_i^U$, response sign $s_i^X$, terminal pre-origin impulse $t_0$ with multiplier $\gamma_0$, and a one-sided response kernel.  With delay $\delta$ and time-scale parameter $h$ (the half-life for the exponential family), the raw kernel is
\begin{equation}
 \bar k_{i8}(r)=
 \begin{cases}
 0, & r<\delta,\\
 \exp[-(\log 2)(r-\delta)/h], & \text{exponential},\\
 \dfrac{r-\delta}{h}\exp[1-(r-\delta)/h], & \text{delayed gamma},\\
 \max\!\left(0,1-\dfrac{r-\delta}{2h}\right), & \text{triangular},
 \end{cases}
 \label{eq:app-sf8-raw-kernel}
\end{equation}
The peak-normalized kernel is
\begin{equation}
 k_{i8}(r)=\frac{\bar k_{i8}(r)}{\max_{v\ge0}\bar k_{i8}(v)},
 \qquad \max_{r\ge0}k_{i8}(r)=1.
 \label{eq:app-sf8-kernel}
\end{equation}
Define the signed impulse path
\begin{equation}
 I_{i8}(u)=s_i^U
 \left[
 \sum_{t\in\mathcal E_{i8}}\mathbf 1\{u=t\}
 +\gamma_0\mathbf 1\{u=t_0\}
 \right].
 \label{eq:app-sf8-impulse-path}
\end{equation}
Let $\sigma^U_{iq}$ be the source covariate scale.  The raw covariate and target components and their shared setting scaling are
\begin{equation}
 \begin{aligned}
 G^U_{i8}(u,j)
 &=\mathbf 1\{j=q\}\,\sigma^U_{iq}I_{i8}(u),\\
 G^X_{i8}(u,j)
 &=\mathbf 1\{j=d\}\,\sigma_{id}s_i^X
   \sum_{v\le u}I_{i8}(v)k_{i8}(u-v),\\
 (\Delta X_{i8\ell},\Delta U_{i8\ell})
 &=\frac{\lambda_{i8\ell}}{q_{i8}}(G^X_{i8},G^U_{i8}).
 \end{aligned}
 \label{eq:app-sf8-covariate-response}
\end{equation}
Repeated historical impulses and the terminal impulse use the same kernel.  A future covariate impulse is included only when the task declares that covariate known over the horizon; past-only covariates stop at the forecast origin.  SF8 constructs at most eight candidate structural configurations; after feasibility qualification, the seed selects one qualified structural configuration, and $\lambda_{i8\ell}$ jointly scales the covariate impulse and target response without changing their signs, timing, target, covariate, or kernel.

\section{Inspection, Sifting, and Evaluation Protocol}
\label{app:protocol}

This appendix gives the admission rules, execution contracts, and metric definitions for the constructions in Appendix~\ref{app:capability-mechanisms}. Empirical support is reported in Appendix~\ref{app:capability-results}.

\subsection{Inspection and Sifting}
\label{app:verification}

\paragraph{Check scopes and experiment binding.}
Structural Feature Expectation, Source Distance, and physical or task checks operate on individual samples; Distribution Difference and Diversity operate on the generated sample set relative to a declared source population. The contracts below specify both qualification scopes. A reported run binds each check to its descriptor, population, statistic, calibrated threshold or coverage requirement, and observed decision.

\subsubsection{Sample Inspection}

Sample Inspection evaluates each realized generated sample $V_{is\ell}$ against four recorded requirements: consistency with the structure selected in Augmentation; the realized Structural Feature Expectation and non-target tolerance; Source Distance and physical or task constraints on every required model-visible history suffix; and any feature-specific horizon or future-effect support condition.  The Source Distance evidence uses Equation~\eqref{eq:app-source-distance}: every required history view must have macro distance in $[0.10,2.0]$, and every affected channel must have distance at most $3.0$.  The verification record stores pass or nonnumeric failure states, and numeric aggregation includes scored samples with their recorded support.

\subsubsection{Admitting a complete sample set}

A candidate sample set $\widetilde{\mathcal V}_{is}$ qualifies by satisfying all of the following conditions:
\begin{enumerate}[leftmargin=*]
  \item Structural Feature $s$ is applicable to $R_i'$, a generated sample is present for every $\ell\in\mathcal L_s$, and every sample passes Sample Inspection;
  \item all generated samples share the selected Reference $R_i'$, fields, and structural configuration required by Equations~\eqref{eq:app-generated-sample-set} and~\eqref{eq:app-field-level-transformation}; and
  \item the ordered Control Settings and other sample-set construction invariants of Structural Feature $s$ hold across the complete set; and
  \item the set-level Structural Feature Descriptor distribution satisfies the declared Distribution Difference bound and its structural configurations and Control Settings satisfy the declared Diversity requirement.
\end{enumerate}
Each $(R_i',s)$ pair contributes all $|\mathcal L_s|$ generated samples or a recorded nonnumeric state, while its original Reference remains in $\mathcal A_t$. Denote a passing set by $\mathcal V_{is}$; the qualified sample sets of task $t$ form
\begin{equation}
 \mathcal P_t=\{(R_i',\mathcal V_{is}):\widetilde{\mathcal V}_{is}\text{ passes sample-set qualification}\}.
 \label{eq:app-qualified-sample-set}
\end{equation}
The record stores the reason and non-numeric state for each inapplicable or rejected construction.  Structural Features that declare a minimum future change enforce it during sample-set construction; other features apply truth-change sufficiency at NRMSE scoring, while an eligible MASE value remains available.

\paragraph{Task-level contract verification.} Verification checks treatment records for source distance, mechanism support, and applicable horizon support, recording accepted and rejected contracts. When the source data support exact replay, it also reconstructs the References and generated samples and compares their task records. Write $\operatorname{VerOK}_t=1$ when the declared policy accepts the task.

\paragraph{Verified evaluation bundle.}

For an accepted task, Stage~3 assembles
\begin{equation}
 \mathcal D_t=\left(\mathcal A_t,\mathcal P_t\right)
 \quad\text{if }\operatorname{VerOK}_t=1.
 \label{eq:app-evaluation-bundle}
\end{equation}
The bundle carries the links between each Reference and generated sample together with the accepted verification policy and report.  Inference receives only bundles with an accepted report.  Any SF6--SF8 ablated-input view is derived from an admitted sample after verification and recorded under a separate inference-ablation contract.

\subsection{Evaluation}
\label{app:profile-table}

\subsubsection{Model Configuration and Execution States}
\label{app:native-inference}

Each paired Reference--sample comparison uses one fixed model configuration, including model identity, preprocessing, target/covariate mapping, context policy, and permitted adaptation. Auxiliary-input pairs additionally fix the randomization policy and change only the declared input view.

\paragraph{Input projection and output restoration.} The Evaluator checks horizon and input compatibility, projects each constructed object to the model context, restores target order after any permitted channel-wise adaptation, and records unsupported, failed, and nonfinite requests separately from valid forecasts. Auxiliary removal is evaluated only for configurations that receive the designated input.

Metric scoreability is recorded separately from execution: each metric uses only valid forecasts and required pairings, while unsupported and nonnumeric states retain their coverage counts.

\subsubsection{MASE scale from the Reference history}
\label{app:mase}

Mean absolute scaled error (MASE)~\cite{hyndman2006another} measures forecast error across target scales.  Let $L_i$ be the history length, let $p_t$ be the adapter-resolved seasonality, and set $\tau_i=\min(\max(1,p_t),\max(1,L_i-1))$.  For target $d$, let $\mathcal Q_{id}(\tau_i)$ contain the history-index pairs admitted by the adapter-resolved scale policy.  A complete repaired-history policy admits the repaired cells, whereas a history-observation-preserving scale policy admits pairs whose two source cells are observed.  The Reference-history scale is
\begin{equation}
 \sigma^{\mathrm{MASE}}_{id}=\frac{1}{|\mathcal Q_{id}(\tau_i)|}\sum_{u\in\mathcal Q_{id}(\tau_i)}\left|H_{i,u,d}-H_{i,u-\tau_i,d}\right|.
 \label{eq:app-mase-scale}
\end{equation}
The evaluator uses the declared fallback when the seasonal-pair scale is unavailable or degenerate. The resulting Reference-history scale is shared by the Reference, generated samples, and any eligible ablation pair. 

For each Reference and SF, its generated samples and any eligible full/ablated inference pair share the resulting $\sigma^{\mathrm{MASE}}_{id}$.

For object type $X\in\{R,V\}$, let $\Omega_i^X$ contain its observed future cells.  Its instance-level score is
\begin{equation}
 \operatorname{MASE}_i^X=\frac{1}{|\Omega_i^X|}\sum_{(h,d)\in\Omega_i^X}\frac{|\widehat Y_{i,h,d}^{X}-Y_{i,h,d}^{X}|}{\sigma^{\mathrm{MASE}}_{id}}.
 \label{eq:app-instance-mase}
\end{equation}
Within a fixed task and result unit, the evaluator averages instance-level values equally.  Lower MASE therefore denotes smaller forecast error relative to the Reference history's variation scale.  For $X=R$, this result object is Reference MASE; for $X=V$, it is Probe MASE.  The input-ablation comparison uses the same formula restricted to the assessed target under the full and ablated input views.

\subsubsection{NRMSE for response to controlled changes}
\label{app:statistics}

Fix task $t$, model configuration $m$, Structural Feature $s$, and control setting $\ell$.  For an affected and observed future cell $k=(h,d)$ of instance $i$, let superscript $(a)$ denote the generated-sample view and superscript $(0)$ denote its selected Reference.  Define the truth and forecast changes as
\begin{equation}
 \Delta Y_{ik}=Y^{(a)}_{ik}-Y^{(0)}_{ik},
 \qquad
 \Delta\widehat Y_{ik}
 =\widehat Y^{(a)}_{ik}-\widehat Y^{(0)}_{ik}.
 \label{eq:app-forecast-changes}
\end{equation}
Let $\Omega_i^{\Delta}$ contain the affected observed cells declared by $T_s$, and use the Reference-history scale $\sigma^{\mathrm{MASE}}_{ik}=\sigma^{\mathrm{MASE}}_{id}$.  The standardized truth-change root-mean-square is
\begin{equation}
 q_i
 =\sqrt{
 \frac{1}{|\Omega_i^{\Delta}|}
 \sum_{k\in\Omega_i^{\Delta}}
 \left(\frac{\Delta Y_{ik}}{\sigma^{\mathrm{MASE}}_{ik}}\right)^2
 }.
 \label{eq:app-truth-change-root-mean-square}
\end{equation}
The eligible set $\mathcal I^{\Delta}_{t,m,s,\ell}$ contains instances with two valid forecasts, nonempty $\Omega_i^{\Delta}$, and $q_i\ge 0.05$.  The task-level normalized root-mean-square error (NRMSE) is
\begin{equation}
 \operatorname{NRMSE}_{t,m,s,\ell}
 =\sqrt{
 \frac{
 \displaystyle
 \sum_{i\in\mathcal I^{\Delta}_{t,m,s,\ell}}
 \sum_{k\in\Omega_i^{\Delta}}
 \left(
   \frac{\Delta\widehat Y_{ik}-\Delta Y_{ik}}{\sigma^{\mathrm{MASE}}_{ik}}
 \right)^2
 }{
 \displaystyle
 \sum_{i\in\mathcal I^{\Delta}_{t,m,s,\ell}}
 \sum_{k\in\Omega_i^{\Delta}}
 \left(
   \frac{\Delta Y_{ik}}{\sigma^{\mathrm{MASE}}_{ik}}
 \right)^2
 }}.
 \label{eq:app-nrmse}
\end{equation}
The ratio pools standardized squared sums over eligible instances and cells and requires a nonempty set with positive truth-change energy. Equivalently, its square is the mean of the squared pair-level ratios weighted by each pair's standardized target-change energy, which includes its number of affected observed cells.  Instances below the $0.05$ truth-change threshold remain available to MASE and retain a non-numeric NRMSE state.

\subsubsection{Auxiliary-Input Gap after removing an auxiliary input}

The inference ablation applies to model/view pairs that expose the designated auxiliary input.  For the SF6 intervention, the highest-absolute-loading target is assessed and the ablated view retains only that target, deleting every other target channel.  For the SF7 intervention, the responder is assessed and the ablated view retains only the responder, again deleting every other target channel.  For the SF8 intervention, the target panel is retained in both views while the ablated view deletes the designated dynamic-covariate carrier together with its corresponding name, availability, future-visibility, and type metadata.  For both SF6 and SF7, task-permitted non-target covariates remain available in the full and ablated views. The SF6/SF7 comparison includes configurations receiving other target channels; the SF8 comparison includes configurations receiving the designated carrier.

Let $\Omega_i^{\mathrm{abl}}$ contain the observed future cells of the feature-specific assessed target.  Let $A_i^{(a)}$ and $A_i^{\mathrm{abl}}$ denote Equation~\eqref{eq:app-instance-mase} restricted to $\Omega_i^{\mathrm{abl}}$ for the full generated-sample and ablated-input views, respectively.  The two views share the assessed-target history and future, forecast origin, horizon, observation mask, and Reference-history scale.  Let $\mathcal I^{\mathrm{abl}}_{t,m,s,\ell}$ contain the eligible pairs for which both forecasts are scoreable.  Auxiliary-Input Gap is
\begin{equation}
 \Delta\operatorname{MASE}_{t,m,s,\ell}
 =\frac{1}{|\mathcal I^{\mathrm{abl}}_{t,m,s,\ell}|}
 \sum_{i\in\mathcal I^{\mathrm{abl}}_{t,m,s,\ell}}
 \left(
   A_i^{\mathrm{abl}}
   -A_i^{(a)}
 \right).
 \label{eq:app-input-sensitivity}
\end{equation}
Positive values mean that auxiliary-input removal increases assessed-target error, zero denotes no average signed difference, and negative values mean that removal decreases the error.  The quantity records net predictive contribution under the declared deletion.

\subsubsection{Recorded Analysis Aggregation and Uncertainty}
\label{app:missingness}

The analyses in this paper use the following \textsc{CaFE}-native task-equal rule; source-benchmark evaluator and leaderboard parity are separate validation targets.  For a result unit $r$ selected by the analysis contract for suite-level reporting, let $\operatorname{Metric}_{t,r,m}$ be the task-level value for model configuration $m$, and let $\mathcal T_{r,m}$ be the tasks on which that value is computed.  The unit identifies one metric and its evaluation object or comparison condition, plus a structural feature and control setting where applicable.  Its recorded task-equal suite result is
\begin{equation}
 \operatorname{SuiteMetric}_{r,m}
 =\frac{1}{|\mathcal T_{r,m}|}
 \sum_{t\in\mathcal T_{r,m}}
 \operatorname{Metric}_{t,r,m},
 \qquad |\mathcal T_{r,m}|>0.
 \label{eq:app-task-equal-aggregation}
\end{equation}
Thus, each benchmark task contributes one value regardless of its number of References or qualified sample sets. The suite row records the task-set cardinality, and task-level values remain available for paired comparisons.

For paired model comparisons, the task set is $\mathcal T_{r,m_1}\cap\mathcal T_{r,m_2}$, and the analysis averages one difference per common task with equal weight. The bootstrap resamples tasks for a single-model result and paired task differences for a model comparison; the analysis contract records its seed, repetition count, interval rule, and task identities. Each metric retains its own scoreable support. The secondary structural-response macro uses equally weighted, direction-corrected SF-local ranks on common model support and excludes raw NRMSE pooling and Auxiliary-Input Gap.

\subsection{Evaluation and Replay Scope}
\label{app:replay-scope}

\paragraph{Evaluation records.}
Each result unit retains its benchmark task, Reference, structural feature, control setting, model configuration, and scoreable support. These fields keep paired comparisons aligned while allowing unsupported or nonnumeric cases to remain explicit:
\begin{equation}
 \mathfrak S_t
 \xrightarrow{\mathrm{Stages\ 1\text{--}2}}
 \mathfrak G_t
 \xrightarrow{\mathrm{Stage\ 3}}
 \left(\mathcal D_t,\mathfrak V_t\right)
 \xrightarrow{m}
 \mathfrak P_{t,m}
 \xrightarrow{\mathrm{Metrics}}
 \mathfrak A_{t,m},
 \label{eq:app-evaluation-chain}
\end{equation}
where $\mathfrak S_t$ denotes the benchmark-task input, $\mathfrak G_t$ the generation conditions, $\mathfrak V_t$ the qualification result, $\mathfrak P_{t,m}$ the predictions, and $\mathfrak A_{t,m}$ the metric results. The analysis records each result unit's task-set cardinality and keeps the task-level values available for paired comparisons.

The retained records support progressively broader checks, from treatment verification to construction and metric reconstruction and then end-to-end forecasting replay. Appendix~\ref{app:reproduction-guide} specifies the required inputs and procedures for each reproduction level.

\section{Supporting Results and Reproducibility}
\label{app:capability-results}

The following sections supply experiment-specific support, aggregation details, and complete numerical results in the order of the main experiments. Earlier analysis blocks retain their original populations; reproduction requirements follow the results.

\subsection{Rank-Correlation Definitions}
\label{app:rank-correlation}

For two quantities $X$ and $Y$ observed on the same $n$ items, Spearman's rank correlation is the Pearson correlation of their ranks, $\rho_s=\operatorname{corr}(R_X,R_Y)$; when there are no ties, this is equivalently $\rho_s=1-6\sum_{i=1}^{n}d_i^2/[n(n^2-1)]$, where $d_i$ is the difference between the two ranks~\cite{spearman1904rank}. We use average ranks for tied values.

Kendall's tie-adjusted rank correlation is reported as
\begin{equation}
\tau_b=\frac{C-D}{\sqrt{(C+D+T_X)(C+D+T_Y)}},
\label{eq:app-kendall-tau-b}
\end{equation}
where $C$ and $D$ count concordant and discordant item pairs, and $T_X$ and $T_Y$ count pairs tied only in $X$ and only in $Y$, respectively~\cite{kendall1938rank}. Thus $\rho_s$ summarizes agreement after rank transformation, whereas $\tau_b$ measures pairwise ordering agreement while accounting for ties; both statistics are computed on the stated common model support.

\subsection{Computational Environment}

Experiments used a standard KVM-based compute-node configuration with 48 Intel Xeon Gold 6530 vCPUs, 188~GiB of system memory, and four NVIDIA GeForce RTX 5090 GPUs with 32,607~MiB of memory per GPU, running Ubuntu 24.04.4 LTS. The \textsc{CaFE} data-generation, validation, and aggregation pipeline used Python 3.12.3. Model inference services used Python 3.11.15, PyTorch 2.10.0, CUDA 12.8, cuDNN 9.10, and NVIDIA driver 575.57.08. This profile establishes the standard per-node hardware and software environment. Appendix~\ref{app:evaluation-coverage} summarizes the corresponding evaluation support and analysis scope.

\subsection{Evaluation Support and Aggregation}

Table~\ref{tab:app-current-main-support} records the supported horizons, configuration counts, and model roster for each frozen suite. Timer-3.5 denotes the evaluated Timer-S1 checkpoint~\cite{thuml2026timers1}.

\begin{table}[t]
  \centering
  \caption{Displayed model support of the four latest result packages. Short and FEV use 11 models; Medium and Long use 10 because TiREx-2 is outside the supported horizon.}
  \label{tab:app-current-main-support}
  \scriptsize
  \setlength{\tabcolsep}{3.6pt}
  \begin{tabular}{@{}lrrr@{}}
    \toprule
    Suite & Horizon & Configurations & Models \\
    \midrule
    GIFT-Eval Short   & 30--60  & 55 & 11 \\
    GIFT-Eval Medium  & 480--600 & 21 & 10 \\
    GIFT-Eval Long    & 720--900 & 21 & 10 \\
    fev-bench Mini-20   & 5--288   & 20 & 11 \\
    \bottomrule
  \end{tabular}
\end{table}

Suite aggregation follows Equation~\eqref{eq:app-task-equal-aggregation}. Probe MASE and Forecast-Change NRMSE first average eligible task-level values within each SF--setting cell, then apply equal setting weights within the feature; NRMSE pools standardized squared terms within each task as in Equation~\eqref{eq:app-nrmse}. Feature-local rankings use common support and average ranks for ties. The secondary macro maps direction-corrected ranks to $[0,1]$ before averaging SF1--SF8 equally; Auxiliary-Input Gap is excluded.

\ifdefined\CaFEStandaloneAppendix
\begin{figure}[t]
  \centering
  \IfFileExists{pics/fig_cafe_main_capability_mase_heatmaps.pdf}{%
    \includegraphics[clip,width=0.98\textwidth]{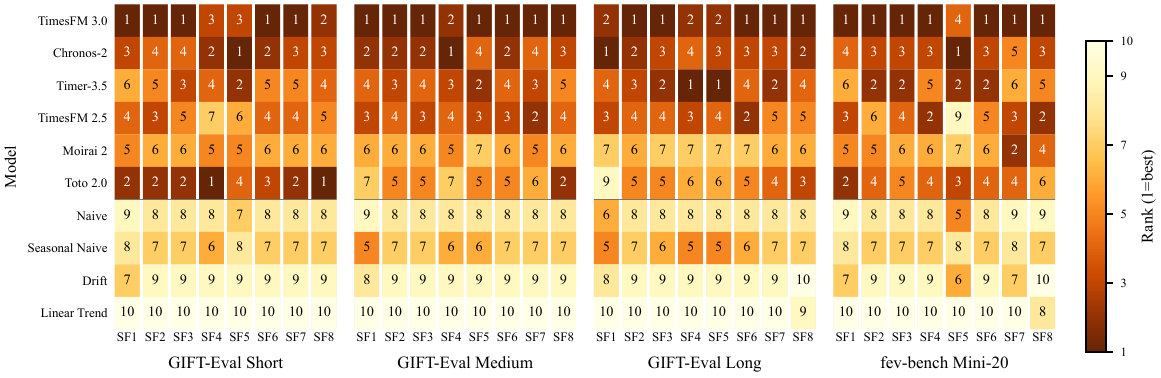}%
  }{\fbox{\parbox[c][0.29\textwidth][c]{0.92\textwidth}{\centering structural-feature Probe MASE heatmaps}}}
  \caption{Probe MASE by suite, model, and structural feature for the four latest panels in one horizontal strip. Each cell preserves the panel's source-window, unordered-variant, and configuration aggregation; model labels appear once at the left.}
  \label{fig:main-capability-mase-heatmaps}
\end{figure}

\begin{figure}[t]
  \centering
  \IfFileExists{pics/fig_cafe_main_capability_effect_heatmaps.pdf}{%
    \includegraphics[clip,width=0.98\textwidth]{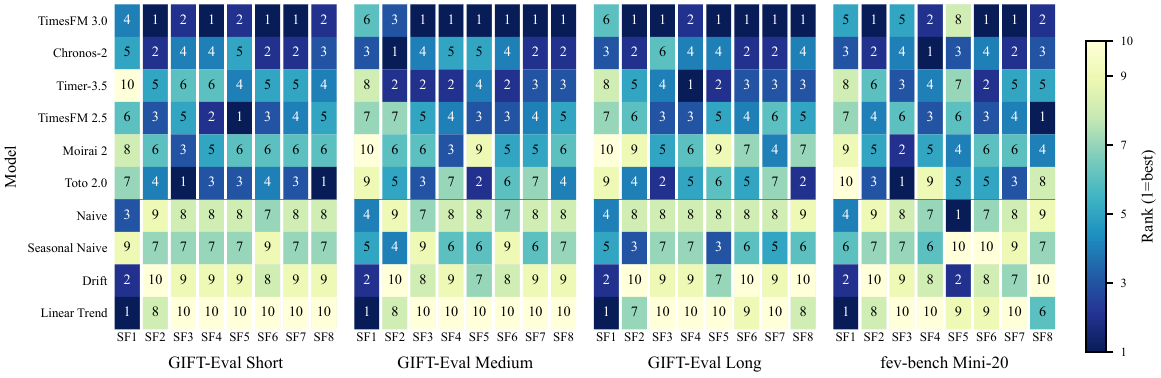}%
  }{\fbox{\parbox[c][0.29\textwidth][c]{0.92\textwidth}{\centering structural-feature Forecast-Change NRMSE heatmaps}}}
  \caption{Forecast-Change NRMSE by suite, model, and structural feature for the four latest panels in one horizontal strip. Lower is better within each feature and its scoreable support; numeric cells are primary and the shared log1p color scale is descriptive.}
  \label{fig:main-capability-effect-heatmaps}
\end{figure}
\fi

\begin{table}[t]
\centering
\caption{Secondary equal-weight structural-response ranking on common model support. Each SF score is converted to a direction-corrected feature-local rank score before averaging equally across SF1--SF8; raw NRMSE values are not pooled.}
\label{tab:latest-four-suite-macro}
\scriptsize
\begin{tabularx}{\textwidth}{@{}lX@{}}
\toprule
Suite & Equal-weight diagnostic ranking \\
\midrule
    GIFT-Eval Short & 1. TimesFM 3.0 (0.925); 2. Toto 2.0 (0.625); 3. Chronos-2 (0.600); 4. TimesFM 2.5 (0.550); 5. Timer-3.5 (0.175); 6. Moirai 2 (0.125) \\
    GIFT-Eval Medium & 1. TimesFM 3.0 (0.925); 2. Timer-3.5 (0.650); 3. Chronos-2 (0.600); 4. TimesFM 2.5 (0.375); 5. Toto 2.0 (0.300); 6. Moirai 2 (0.150) \\
    GIFT-Eval Long & 1. TimesFM 3.0 (0.950); 2. Chronos-2 (0.600); 3. Timer-3.5 (0.600); 4. TimesFM 2.5 (0.400); 5. Toto 2.0 (0.375); 6. Moirai 2 (0.075) \\
    fev-bench Mini-20 & 1. Chronos-2 (0.750); 2. TimesFM 3.0 (0.700); 3. TimesFM 2.5 (0.500); 4. Toto 2.0 (0.375); 5. Timer-3.5 (0.350); 6. Moirai 2 (0.325) \\
\bottomrule
\end{tabularx}
\end{table}

\subsection{Accuracy and Response Fidelity}

The winner table uses the complete evaluated roster, whereas the common ten-model heatmap display omits TiREx-2. The two views therefore compare different model sets on Short and fev-bench Mini-20.

\subsubsection{Latest four-panel inventory and winners}
\label{app:latest-four-suite}

Table~\ref{tab:latest-four-suite-support} records the panel prediction counts. These panels retain the source-window, unordered-variant, and configuration aggregation supplied by their numerical inputs.

\begin{table}[t]
\centering
\caption{Inventory of the four latest result panels after restricting each view to the comparison roster.}
\label{tab:latest-four-suite-support}
\scriptsize
\begin{tabularx}{\textwidth}{@{}lrrrr@{}}
\toprule
Suite & Configs & Models shown & Panel predictions \\
\midrule
    GIFT-Eval Short & 55 & 11 & 399,740 \\
    GIFT-Eval Medium & 21 & 10 & 127,380 \\
    GIFT-Eval Long & 21 & 10 & 121,810 \\
    fev-bench Mini-20 & 20 & 11 & 121,902 \\
\bottomrule
\end{tabularx}
\end{table}
\begin{table}[t]
\centering
\caption{Winners for each structural feature in the four latest result panels. Entries give the model and the number of feature tracks won; ties would be reported jointly.}
\label{tab:latest-four-suite-winners}
\scriptsize
\begin{tabularx}{\textwidth}{@{}lXX@{}}
\toprule
Suite & Lowest Probe MASE & Lowest Forecast-Change NRMSE \\
\midrule
    GIFT-Eval Short & TimesFM 3.0 (5/8); Toto 2.0 (2/8); Chronos-2 (1/8) & TimesFM 3.0 (4/8); Toto 2.0 (2/8); Linear Trend (1/8); TimesFM 2.5 (1/8) \\
    GIFT-Eval Medium & TimesFM 3.0 (7/8); Chronos-2 (1/8) & TimesFM 3.0 (6/8); Chronos-2 (1/8); Linear Trend (1/8) \\
    GIFT-Eval Long & TimesFM 3.0 (5/8); Timer-3.5 (2/8); Chronos-2 (1/8) & TimesFM 3.0 (6/8); Linear Trend (1/8); Timer-3.5 (1/8) \\
    fev-bench Mini-20 & TimesFM 3.0 (7/8); TiREx-2 (1/8) & TimesFM 3.0 (3/8); Chronos-2 (1/8); Linear Trend (1/8); Naive (1/8); TimesFM 2.5 (1/8); Toto 2.0 (1/8) \\
\bottomrule
\end{tabularx}
\end{table}

Figures~\ref{fig:main-capability-mase-heatmaps} and~\ref{fig:main-capability-effect-heatmaps} display the cell values. The main-paper radar restricts the NRMSE view to six models shared by all suites and uses a separate scale for each feature.

\subsection{Intrinsic Model Selection}
\label{app:model-shortlisting}

The regime-switching coverage table below belongs to an earlier frozen score-cache analysis. It retains that analysis's seven-family slice, separately from the current intrinsic-selection result in the main paper.

The model-shortlisting analysis uses a frozen score-cache population distinct from the four latest suite panels. The cache contains 55 configurations, 19 source families, 15,673 common-scored original windows, and 6,016 capability--window rows; a capability row can correspond to the same original window as another capability. Twelve model columns are retained, while the recommendation analysis uses eight frozen foundation-model configurations. The analysis record retains their names and complete input order.

For the leave-source-family-out analysis, the held-out family is excluded from all A-side rankings and threshold fitting. The generic strategy uses all A-side original windows, the qualified strategy uses capability-qualified A-side original windows, and the augmented strategy averages the matched augmentation variants within each window and ranks by augmented MASE. Within each family, windows are averaged by configuration before equal-family Borda ranks are averaged across families. The A-side Q75 strength cutoff is applied to B; B is scored only on original windows and the family-balanced champion is determined after equal configuration aggregation. The effect-NRMSE ranking is reported as a separate response diagnostic; MASE determines the recommendations.

The pooled case uses the underlying collection plus original item as the source-group key. All windows, frequency variants, and augmentation variants for one group remain on the same side. Five seeds and both directions yield ten A/B directions; Q75 is fitted on A and applied unchanged to B. The reported Bitbrains subset is an exploratory post-hoc slice selected after inspection of the completed grid and is evaluated only on original MASE. The ten directions share a source pool and are summarized descriptively; no significance test is applied, so this slice provides exploratory descriptive evidence. The A-only restriction applies to threshold fitting and ranking; upstream qualification retains its offline future-quality and feasibility checks.

\begin{table}[t]
  \centering
  \caption{Regime-switching candidate coverage in the leave-source-family-out analysis. A hit means that the B-side family-balanced original-MASE champion is present in the frozen Top-$K$ list.}
  \label{tab:model-shortlisting-coverage}
  \scriptsize
  \setlength{\tabcolsep}{6pt}
  \begin{tabular}{@{}lrrr@{}}
    \toprule
    Recommendation & Top-1 & Top-3 & Top-5 \\
    \midrule
    Generic original MASE & 1/7 & 3/7 & 7/7 \\
    Qualified original MASE & 1/7 & 4/7 & 7/7 \\
    Capability-augmented MASE & 1/7 & 6/7 & 7/7 \\
    \bottomrule
  \end{tabular}
\end{table}

\begin{table}[t]
  \centering
  \caption{Bitbrains exploratory case under Q75 family-Borda selection. Values are direction-equal descriptive means over ten disjoint source-group directions; the B-side support has at least 16 A groups and 14 B groups per direction.}
  \label{tab:model-shortlisting-bitbrains}
  \scriptsize
  \setlength{\tabcolsep}{5pt}
  \begin{tabular}{@{}lrr@{}}
    \toprule
    Recommendation & Mean gain vs generic & Wins--ties--losses \\
    \midrule
    Qualified original MASE & 11.02\% & -- \\
    Capability-augmented MASE & 9.06\% & 8--1--1 \\
    \bottomrule
  \end{tabular}
\end{table}


\subsection{Relational Feature Validation}

The historical numerical block retains the input-removal analysis's task counts and descriptive intervals. Its population is separate from the latest suite panels; the deletion and assessed-target contracts are specified in Appendix~\ref{app:protocol}.

\subsection{Augmentation Faithfulness: Paired-Ranking Study}
\label{app:small-model-protocol}

The small-model paired-ranking study contains five architectures, eight native GIFT-Eval configurations, three training seeds, and three augmentation seeds; the seven analyzed capabilities are trend, multi-seasonal, time-varying seasonality, regime switching, intermittency, common factor, and cross-series dependence. A is the sampled original-source union (2,294 windows), B is the per-capability subset of original windows whose paired treatment enters C, and C is the valid treatment set. Across capabilities there are 2,712 source pairs; the three augmentation seeds retain 2,468, 2,466, and 2,460 B pairs, with 2,436 common valid pairs and 21,793 valid treatment samples across the three rounds. These counts define the analyzed support for the paired-ranking result.

For each source, unordered treatment variants are averaged first, followed by equal source averaging within each configuration--feature cell, equal averaging over valid configurations within each feature, and equal averaging over the seven feature summaries; absent features are not imputed. B uses the same source membership, target mask, weights, and original full-history seasonal MASE denominator as C, rather than a denominator recomputed from the augmented history or the 512-point model context. Training-seed scores are averaged after the three augmentation seeds; the reported $\pm$ values are sample SD over the three training seeds.

Training uses only the original train/validation splits with context 512 and horizon 48. Minimum original-validation MASE selects the checkpoint across eligible epochs and at most one prescribed regularized retry; selected weights are frozen before testing. Eight model trajectories used a retry and eight used a bounded refinement, while four selected trajectories met the late-overfit diagnostic and therefore used earlier validation-best checkpoints. The analysis reports selected-checkpoint behavior for this suite.

The rank uncertainty analysis uses 1,000 paired bootstrap replicates with fixed selected-suite item clusters, or circular blocks of four consecutive forecast origins for single-item configurations. Training and augmentation seeds are averaged before resampling. The conditional intervals are $\rho\in[0.70,0.90]$ and $\tau_b\in[0.60,0.80]$; the near-first C difference between iTransformer and DLinear has interval $[-0.0269,0.0114]$, which spans zero under this resampling. Support is heterogeneous: the multivariate group has three native items, while common-factor and cross-series dependence have 59 and 39 source windows concentrated in three and two items, respectively. These counts define the scope of the selected-suite result. Effect NRMSE remains a separate response diagnostic, while original MASE supplies the ranking metric.

\begin{table}[t]
  \centering
  \caption{Small-model B-to-C rank agreement by input group and structural feature. Correlations use five architectures; the multivariate group has only three native items.}
  \label{tab:small-model-local-ranks}
  \scriptsize
  \begin{tabular}{@{}lr@{}}
    \toprule
    Group or structural feature & Spearman $\rho$ \\
    \midrule
    Multivariate & 1.00 \\
    Univariate & 0.30 \\
    Common factor & 1.00 \\
    Cross-series dependence & 1.00 \\
    Intermittency & 1.00 \\
    Regime switching & 1.00 \\
    Multi-seasonal & 0.90 \\
    Time-varying seasonality & 0.90 \\
    Trend & $-0.10$ \\
    \bottomrule
  \end{tabular}
\end{table}

Table~\ref{tab:small-model-local-ranks} preserves the input-group and feature-level results; six of the seven features have $\rho\geq0.90$. Table~\ref{tab:small-model-trend-sensitivity} localizes selected trend changes, while the causes of the rank reversals remain unresolved.

\begin{table}[t]
  \centering
  \caption{Trend-treatment sensitivity retained in the small-model package. B is the paired original MASE and C is the valid treatment MASE; lower is better.}
  \label{tab:small-model-trend-sensitivity}
  \scriptsize
  \setlength{\tabcolsep}{5pt}
  \begin{tabular}{@{}lrr@{}}
    \toprule
    Configuration--model & B & C \\
    \midrule
    M4 Hourly--Transformer & 1.8348 & 9.3486 \\
    M4 Hourly--DLinear & 3.0342 & 1.8950 \\
    M4 Hourly--PatchTST & 1.4468 & 2.1678 \\
    SZ Taxi--PatchTST & 0.6568 & 1.4263 \\
    SZ Taxi--TSMixer & 0.9936 & 0.6860 \\
    \bottomrule
  \end{tabular}
\end{table}

\subsection{Stability under Benchmark Refresh}

The ten-seed local-winner table below belongs to an earlier fixed ten-task panel. The main paper's five-seed analysis uses the current 55-configuration Short panel and seven-model roster; the two seed studies retain separate supports.

\subsection{Trade-off between Accuracy and Response Fidelity}

The full checkpoint trajectories are retained below. The matched training conditions and checkpoint schedule are specified under Reproduction Levels and Procedures.

\subsection{Historical and Expanded Numerical Blocks}
\label{app:complete-numerical-data}
\label{app:gift-short-full55}

The expanded Short results and the earlier six-model cross-suite results use separate populations and aggregation rules. The earlier ten-seed stability and input-removal tables retain their original support; the fine-tuning trajectories describe one training realization.

The expanded Short block pairs Linear Trend's SF1 response NRMSE of $3.24\times10^{-6}$ with Probe MASE of 8.074. Its input-removal estimates include an SF6 gap of 0.157 for TimesFM 3.0 and $-0.020$ for Chronos-2, whose 95\% interval includes zero.

\subsection{Earlier Six-Model Cross-Suite Data by Structural Feature}

The expanded GIFT-Eval Short package is reported separately from the earlier six-model cross-suite block. The compact cross-suite tables below report Probe MASE / Forecast-Change NRMSE pairs (P/N); lower is better for each component, and NRMSE $=1$ is the zero-response reference. The expanded Short tables then report Official MASE, matched original / augmented MASE, and Forecast-Change NRMSE in separate blocks. Values are configuration-equal after the package's source-window and unordered-variant reductions.

\begin{table}[t]
  \centering
  \caption{Probe MASE / Forecast-Change NRMSE pairs (P/N) for the structural features on GIFT-Eval Short. Lower is better for each component; NRMSE $=1$ is the zero-response reference.}
  \label{tab:app-capability-data-gift-short}
  \scriptsize
  \setlength{\tabcolsep}{2.8pt}
  \begin{tabular}{@{}lrcccccc@{}}
    \toprule
    SF & $N$ & Chronos-2 & Timer-3.5 & TimesFM 2.5 & TiREx-2 & Moirai 2 & Toto 2.0 \\
    \midrule
    SF1 & 18 & 2.132 / \textbf{0.082} & 1.966 / 0.243 & 1.898 / 0.216 & \textbf{1.843} / 0.256 & 2.152 / 0.215 & 1.876 / 0.243 \\
    SF2 & 20 & 2.370 / 0.779 & 2.758 / 0.946 & 2.499 / 0.802 & 2.532 / 0.844 & 4.171 / 1.010 & \textbf{1.934} / \textbf{0.637} \\
    SF3 & 19 & 1.526 / \textbf{0.529} & 1.738 / 0.674 & \textbf{1.462} / 0.679 & 1.484 / 0.580 & 1.853 / 0.740 & 1.572 / 0.571 \\
    SF4 & 20 & 1.703 / 0.096 & 1.717 / 0.133 & 1.627 / 0.091 & 1.597 / \textbf{0.071} & 1.910 / 0.134 & \textbf{1.574} / 0.144 \\
    SF5 & 20 & 2.929 / 0.905 & 2.872 / 0.914 & 2.833 / \textbf{0.837} & 2.745 / 0.972 & 4.211 / 0.965 & \textbf{2.699} / 0.955 \\
    SF6 & 4 & \textbf{0.618} / \textbf{0.257} & 0.644 / 0.301 & 0.648 / 0.333 & 0.658 / 0.259 & 0.695 / 0.353 & 0.625 / 0.291 \\
    SF7 & 8 & 2.895 / \textbf{0.989} & 2.982 / 1.043 & 2.896 / 1.187 & 2.865 / 1.005 & 3.060 / 1.044 & \textbf{2.746} / 1.460 \\
    SF8 & 6 & 3.638 / 0.626 & 3.833 / 0.554 & 3.491 / \textbf{0.495} & \textbf{3.485} / 0.850 & 5.907 / 0.684 & 3.522 / 0.632 \\
    \bottomrule
  \end{tabular}
\end{table}

\begin{table}[t]
  \centering
  \caption{Expanded GIFT-Eval Short full55 package: Official MASE and matched original / augmented MASE for the first five model configurations.}
  \label{tab:app-gift-short-full55-mase-a}
  \scriptsize
  \setlength{\tabcolsep}{2.1pt}
  \resizebox{\columnwidth}{!}{%
  \begin{tabular}{@{}lrrrrrrrrr@{}}
    \toprule
    Model & Official & SF1 & SF2 & SF3 & SF4 & SF5 & SF6 & SF7 & SF8 \\
    \midrule
    Chronos-2 & 1.567 & 2.208/2.216 & 0.716/0.730 & 1.232/1.241 & 1.575/1.859 & 0.871/0.896 & 0.828/1.194 & 0.773/0.802 & 0.734/0.765 \\
    TimesFM 2.5 & 1.601 & 2.140/2.220 & 0.713/0.722 & 1.244/1.246 & 1.911/2.318 & 0.969/1.052 & 0.828/1.232 & 0.782/0.813 & 0.745/0.782 \\
    TimesFM 3.0 & 1.464 & 2.003/2.050 & 0.667/0.686 & 1.166/1.165 & 1.628/1.916 & 0.885/0.923 & 0.807/1.188 & 0.748/0.772 & 0.697/0.732 \\
    TiREx-2 & 1.610 & 2.239/2.216 & 0.706/0.732 & 1.241/1.246 & 1.722/1.914 & 0.876/0.921 & 0.838/1.233 & 0.788/0.817 & 0.751/0.801 \\
    Moirai 2 & 1.701 & 2.258/2.260 & 0.766/0.797 & 1.353/1.372 & 1.688/1.998 & 0.904/1.030 & 0.922/1.327 & 0.824/0.863 & 0.809/0.877 \\
    \bottomrule
  \end{tabular}}
\end{table}

\begin{table}[t]
  \centering
  \caption{Expanded GIFT-Eval Short full55 panel: Official MASE and matched original / augmented MASE for the remaining six model configurations.}
  \label{tab:app-gift-short-full55-mase-b}
  \scriptsize
  \setlength{\tabcolsep}{2.1pt}
  \resizebox{\columnwidth}{!}{%
  \begin{tabular}{@{}lrrrrrrrrr@{}}
    \toprule
    Model & Official & SF1 & SF2 & SF3 & SF4 & SF5 & SF6 & SF7 & SF8 \\
    \midrule
    Timer-3.5 & 1.757 & 2.534/2.644 & 0.711/0.741 & 1.185/1.182 & 1.644/1.931 & 0.865/0.899 & 0.855/1.260 & 0.797/0.828 & 0.723/0.772 \\
    Toto 2.0 & 1.452 & 1.971/2.065 & 0.673/0.700 & 1.186/1.177 & 1.580/1.848 & 0.934/0.963 & 0.837/1.220 & 0.771/0.793 & 0.690/0.726 \\
    Naive & 2.579 & 3.531/4.370 & 1.543/1.891 & 2.041/2.015 & 2.210/2.595 & 1.064/1.237 & 1.215/1.779 & 1.171/1.239 & 1.678/2.001 \\
    Seasonal Naive & 2.193 & 2.945/3.920 & 1.276/1.549 & 1.722/1.728 & 1.953/2.313 & 1.045/1.254 & 1.143/1.659 & 1.086/1.137 & 1.250/1.335 \\
    Drift & 2.565 & 3.431/3.431 & 1.542/1.897 & 2.060/2.036 & 2.631/3.269 & 1.100/1.283 & 1.248/1.821 & 1.191/1.258 & 1.686/2.012 \\
    Linear Trend & 4.806 & 8.074/8.074 & 2.211/2.449 & 2.566/2.585 & 5.219/7.105 & 1.599/1.867 & 2.155/3.016 & 1.905/2.028 & 1.841/2.168 \\
    \bottomrule
  \end{tabular}}
\end{table}

\begin{table}[t]
  \centering
  \caption{Expanded GIFT-Eval Short full55 package: Forecast-Change NRMSE by structural feature for the first five model configurations.}
  \label{tab:app-gift-short-full55-effect-a}
  \scriptsize
  \setlength{\tabcolsep}{2.1pt}
  \begin{tabular}{@{}lrrrrrrrr@{}}
    \toprule
    Model & SF1 & SF2 & SF3 & SF4 & SF5 & SF6 & SF7 & SF8 \\
    \midrule
    Chronos-2 & 1.371 & 0.437 & 0.655 & 0.401 & 0.991 & 0.392 & 0.427 & 0.420 \\
    TimesFM 2.5 & 1.506 & 0.472 & 0.677 & 0.396 & 0.901 & 0.419 & 0.537 & 0.475 \\
    TimesFM 3.0 & 1.179 & 0.340 & 0.624 & 0.387 & 0.928 & 0.392 & 0.403 & 0.389 \\
    TiREx-2 & 1.119 & 0.534 & 0.650 & 0.403 & 0.941 & 0.418 & 0.497 & 0.504 \\
    Moirai 2 & 2.109 & 0.609 & 0.642 & 0.419 & 1.024 & 0.449 & 0.609 & 0.617 \\
    \bottomrule
  \end{tabular}
\end{table}

\begin{table}[t]
  \centering
  \caption{Expanded GIFT-Eval Short full55 panel: Forecast-Change NRMSE by structural feature for the remaining six model configurations.}
  \label{tab:app-gift-short-full55-effect-b}
  \scriptsize
  \setlength{\tabcolsep}{2.1pt}
  \begin{tabular}{@{}lrrrrrrrr@{}}
    \toprule
    Model & SF1 & SF2 & SF3 & SF4 & SF5 & SF6 & SF7 & SF8 \\
    \midrule
    Timer-3.5 & 2.229 & 0.498 & 0.678 & 0.423 & 0.965 & 0.442 & 0.545 & 0.454 \\
    Toto 2.0 & 1.611 & 0.479 & 0.624 & 0.397 & 0.951 & 0.422 & 0.433 & 0.386 \\
    Naive & 1.000 & 1.156 & 0.750 & 0.560 & 1.091 & 0.554 & 0.650 & 0.789 \\
    Seasonal Naive & 2.213 & 1.020 & 0.747 & 0.532 & 1.068 & 0.563 & 0.628 & 0.760 \\
    Drift & \ensuremath{3.42\!\times\!10^{-6}} & 1.175 & 0.753 & 0.578 & 1.112 & 0.560 & 0.657 & 0.794 \\
    Linear Trend & \ensuremath{3.24\!\times\!10^{-6}} & 1.079 & 1.002 & 0.711 & 1.174 & 0.920 & 1.354 & 0.880 \\
    \bottomrule
  \end{tabular}
\end{table}

\begin{table}[t]
  \centering
  \caption{Expanded GIFT-Eval Short full55 input-removal results. Positive $\Delta\mathrm{MASE}$ favors retaining the auxiliary input; intervals are descriptive task-bootstrap 95\% intervals.}
  \label{tab:app-gift-short-full55-ablation}
  \scriptsize
  \setlength{\tabcolsep}{4.0pt}
  \begin{tabular}{@{}llrrr@{}}
    \toprule
    Model & SF & Tasks & $\Delta\mathrm{MASE}$ & 95\% interval \\
    \midrule
    Chronos-2 & SF6 & 13 & -0.020 & [-0.113, 0.052] \\
    Chronos-2 & SF7 & 12 & 0.000 & [-0.119, 0.113] \\
    Chronos-2 & SF8 & 8 & 0.006 & [0.000, 0.013] \\
    TimesFM 3.0 & SF6 & 13 & 0.157 & [0.064, 0.259] \\
    TimesFM 3.0 & SF7 & 11 & 0.051 & [-0.020, 0.160] \\
    TimesFM 3.0 & SF8 & 7 & 0.003 & [-0.000, 0.009] \\
    Toto 2.0 & SF6 & 13 & -0.022 & [-0.163, 0.081] \\
    Toto 2.0 & SF7 & 12 & 0.033 & [-0.010, 0.107] \\
    \bottomrule
  \end{tabular}
\end{table}

\begin{table}[t]
  \centering
  \caption{Probe MASE / Forecast-Change NRMSE pairs (P/N) for the structural features on GIFT-Eval Medium. Lower is better for each component; NRMSE $=1$ is the zero-response reference.}
  \label{tab:app-capability-data-gift-medium}
  \scriptsize
  \setlength{\tabcolsep}{2.8pt}
  \begin{tabular}{@{}lrcccccc@{}}
    \toprule
    SF & $N$ & Chronos-2 & Timer-3.5 & TimesFM 2.5 & TiREx-2 & Moirai 2 & Toto 2.0 \\
    \midrule
    SF1 & 9 & 2.982 / \textbf{0.051} & \textbf{2.808} / 0.264 & 2.880 / 0.259 & -- & 3.469 / 0.293 & 12.785 / 0.590 \\
    SF2 & 11 & 5.309 / 1.102 & 5.533 / 1.206 & \textbf{4.016} / \textbf{0.961} & -- & 10.671 / 1.729 & 10.488 / 1.131 \\
    SF3 & 10 & 7.051 / \textbf{0.787} & 7.292 / 1.130 & \textbf{3.507} / 0.980 & -- & 10.616 / 1.565 & 9.955 / 1.296 \\
    SF4 & 11 & \textbf{2.378} / \textbf{0.053} & 2.581 / 0.191 & 2.453 / 0.131 & -- & 4.026 / 0.276 & 7.597 / 0.176 \\
    SF5 & 11 & \textbf{2.851} / 1.402 & 3.039 / 1.738 & 2.936 / 1.322 & -- & 3.592 / \textbf{1.115} & 6.459 / 1.556 \\
    SF6 & 4 & \textbf{0.919} / \textbf{0.257} & 0.952 / 0.474 & 0.966 / 0.524 & -- & 1.324 / 0.571 & 1.163 / 0.377 \\
    SF7 & 6 & \textbf{7.690} / \textbf{0.992} & 8.041 / 1.070 & 8.089 / 1.053 & -- & 9.902 / 1.051 & 15.988 / 1.189 \\
    SF8 & 4 & \textbf{5.151} / 0.533 & 6.848 / 0.498 & 6.849 / \textbf{0.373} & -- & 9.773 / 0.457 & 19.072 / 0.550 \\
    \bottomrule
  \end{tabular}
\end{table}

\begin{table}[t]
  \centering
  \caption{Probe MASE / Forecast-Change NRMSE pairs (P/N) for the structural features on GIFT-Eval Long. Lower is better for each component; NRMSE $=1$ is the zero-response reference.}
  \label{tab:app-capability-data-gift-long}
  \scriptsize
  \setlength{\tabcolsep}{2.8pt}
  \begin{tabular}{@{}lrcccccc@{}}
    \toprule
    SF & $N$ & Chronos-2 & Timer-3.5 & TimesFM 2.5 & TiREx-2 & Moirai 2 & Toto 2.0 \\
    \midrule
    SF1 & 10 & 3.597 / \textbf{0.062} & \textbf{3.549} / 0.191 & 3.682 / 0.124 & -- & 4.609 / 0.213 & 16.115 / 0.257 \\
    SF2 & 11 & 6.252 / \textbf{0.914} & 6.780 / 1.168 & \textbf{4.890} / 0.966 & -- & 9.542 / 1.638 & 21.869 / 0.998 \\
    SF3 & 10 & \textbf{4.860} / \textbf{0.859} & 8.021 / 1.223 & 5.632 / 0.961 & -- & 9.717 / 1.190 & 23.432 / 1.119 \\
    SF4 & 11 & \textbf{2.934} / \textbf{0.064} & 3.100 / 0.175 & 3.253 / 0.106 & -- & 4.442 / 0.326 & 16.994 / 0.265 \\
    SF5 & 11 & 3.131 / \textbf{1.051} & \textbf{3.087} / 1.118 & 3.185 / 1.069 & -- & 4.636 / 1.683 & 16.407 / 1.184 \\
    SF6 & 4 & \textbf{1.039} / \textbf{0.385} & 1.144 / 0.739 & 1.113 / 0.772 & -- & 1.749 / 1.084 & 1.324 / 0.663 \\
    SF7 & 6 & \textbf{6.669} / \textbf{0.893} & 7.005 / 0.995 & 7.446 / 1.058 & -- & 10.128 / 1.019 & 36.761 / 1.326 \\
    SF8 & 3 & 9.909 / 0.397 & 13.459 / 0.594 & \textbf{9.512} / \textbf{0.335} & -- & 14.300 / 0.490 & 53.238 / 0.857 \\
    \bottomrule
  \end{tabular}
\end{table}

\begin{table}[t]
  \centering
  \caption{Probe MASE / Forecast-Change NRMSE pairs (P/N) for the structural features on fev-bench Mini-20. Lower is better for each component; NRMSE $=1$ is the zero-response reference.}
  \label{tab:app-capability-data-fev}
  \scriptsize
  \setlength{\tabcolsep}{2.8pt}
  \begin{tabular}{@{}lrcccccc@{}}
    \toprule
    SF & $N$ & Chronos-2 & Timer-3.5 & TimesFM 2.5 & TiREx-2 & Moirai 2 & Toto 2.0 \\
    \midrule
    SF1 & 18 & 1.938 / 0.176 & 1.807 / 0.181 & \textbf{1.469} / 0.111 & 1.736 / 0.130 & 1.625 / \textbf{0.087} & 1.591 / 0.124 \\
    SF2 & 20 & 2.656 / 1.202 & 3.003 / 1.143 & 2.582 / 0.954 & 2.703 / 1.042 & 3.144 / 1.213 & \textbf{1.839} / \textbf{0.809} \\
    SF3 & 20 & 2.190 / 0.812 & 2.140 / 0.756 & 1.744 / 0.734 & 2.040 / 0.794 & 2.045 / 0.797 & \textbf{1.659} / \textbf{0.663} \\
    SF4 & 20 & 1.913 / 0.213 & 1.917 / 0.163 & \textbf{1.562} / 0.120 & 1.853 / 0.136 & 1.647 / \textbf{0.103} & 1.603 / 0.110 \\
    SF5 & 18 & 2.188 / 0.934 & 1.987 / 0.963 & \textbf{1.712} / 0.864 & 2.154 / 0.894 & 2.055 / 1.062 & 1.731 / \textbf{0.852} \\
    SF6 & 9 & 1.766 / \textbf{0.286} & 2.044 / 0.359 & \textbf{1.630} / 0.301 & 1.819 / 0.291 & 1.791 / 0.380 & 1.663 / 0.295 \\
    SF7 & 9 & 1.766 / 0.718 & 2.100 / 0.782 & \textbf{1.678} / 0.763 & 1.693 / \textbf{0.717} & 1.711 / 0.834 & 1.826 / 0.752 \\
    SF8 & 8 & 3.073 / 0.776 & 3.046 / 0.642 & 2.480 / 0.691 & 2.972 / 0.651 & 3.025 / 0.674 & \textbf{2.303} / \textbf{0.551} \\
    \bottomrule
  \end{tabular}
\end{table}

\begin{table}[t]
  \centering
  \caption{Spearman rank correlations among Reference MASE (R), Probe MASE (P), and NRMSE (N) for the displayed model sets; the table reports point correlations, with interval estimation treated separately.}
  \label{tab:app-metric-rank-correlations}
  \scriptsize
  \setlength{\tabcolsep}{4.2pt}
  \begin{tabular}{@{}lrrrr@{}}
    \toprule
    Suite & Models & $\rho_{R,P}$ & $\rho_{R,N}$ & $\rho_{P,N}$ \\
    \midrule
    GIFT-Eval Short & 6 & 0.943 & -0.086 & 0.029 \\
    GIFT-Eval Medium & 5 & 0.700 & 0.800 & 0.800 \\
    GIFT-Eval Long & 5 & 0.900 & 0.800 & 0.900 \\
    fev-bench Mini-20 & 6 & 0.657 & 0.543 & 0.771 \\
    \bottomrule
  \end{tabular}
\end{table}

\begin{table}[t]
  \centering
  \caption{Local-winner stability across ten GIFT-Eval Short seeds for the structural features.  Unanimous settings retain one lowest-NRMSE model in all seeds; modal fractions are descriptive.}
  \label{tab:app-stability-by-capability}
  \scriptsize
  \setlength{\tabcolsep}{4.0pt}
  \begin{tabular}{@{}lrrrl@{}}
    \toprule
    SF & Unanimous / 5 & Mean modal & Min modal & Setting-modal models \\
    \midrule
    SF1 & 5 & 1.00 & 1.00 & Chronos-2 (5) \\
    SF2 & 0 & 0.82 & 0.70 & TimesFM 2.5 (1), Toto 2.0 (4) \\
    SF3 & 5 & 1.00 & 1.00 & Chronos-2 (5) \\
    SF4 & 0 & 0.68 & 0.50 & TiREx-2 (3), TimesFM 2.5 (1), Chronos-2 (1) \\
    SF5 & 0 & 0.74 & 0.50 & TimesFM 2.5 (4), Chronos-2 (1) \\
    SF6 & 0 & 0.64 & 0.50 & Chronos-2 (5) \\
    SF7 & 0 & 0.60 & 0.50 & TiREx-2 (4), Chronos-2 (1) \\
    SF8 & 2 & 0.86 & 0.70 & TimesFM 2.5 (5) \\
    \bottomrule
  \end{tabular}
\end{table}

\begin{table}[t]
  \centering
  \caption{Complete Auxiliary-Input Gap results for the evaluated suite--model--feature cells.  Each estimate is $\mathrm{MASE}_{\mathrm{ablated}}-\mathrm{MASE}_{\mathrm{full}}$ with a descriptive task-bootstrap 95\% interval; positive values favor the full auxiliary input.}
  \label{tab:app-complete-input-sensitivity}
  \scriptsize
  \setlength{\tabcolsep}{4.0pt}
  \providecommand{\suppdataci}[3]{\shortstack{#1\\{\tiny [#2, #3]}}}
  \begin{tabular}{@{}lllrrrr@{}}
    \toprule
    Suite & SF & Model & Tasks & $\Delta$MASE [95\% interval] & Task median & Positive tasks \\
    \midrule
    GIFT-Eval Short & SF6 & Chronos-2 & 4 & \suppdataci{0.038}{0.009}{0.068} & 0.042 & 3/4 \\
    GIFT-Eval Short & SF6 & TiREx-2 & 4 & \suppdataci{-0.004}{-0.006}{-0.001} & -0.005 & 1/4 \\
    GIFT-Eval Short & SF6 & Toto 2.0 & 4 & \suppdataci{0.005}{-0.012}{0.030} & -0.004 & 2/4 \\
    GIFT-Eval Short & SF7 & Chronos-2 & 8 & \suppdataci{0.012}{-0.001}{0.030} & 0.002 & 6/8 \\
    GIFT-Eval Short & SF7 & TiREx-2 & 8 & \suppdataci{0.003}{-0.002}{0.008} & 0.001 & 5/8 \\
    GIFT-Eval Short & SF7 & Toto 2.0 & 8 & \suppdataci{0.031}{-0.002}{0.077} & -0.001 & 4/8 \\
    GIFT-Eval Short & SF8 & Chronos-2 & 6 & \suppdataci{-0.001}{-0.012}{0.010} & 0.001 & 4/6 \\
    GIFT-Eval Short & SF8 & TiREx-2 & 6 & \suppdataci{0.001}{\ensuremath{-1.21\!\times\!10^{-4}}}{0.003} & \ensuremath{2.29\!\times\!10^{-4}} & 4/6 \\
    \addlinespace[2pt]
    GIFT-Eval Medium & SF6 & Chronos-2 & 4 & \suppdataci{-0.007}{-0.082}{0.050} & 0.012 & 3/4 \\
    GIFT-Eval Medium & SF6 & Toto 2.0 & 4 & \suppdataci{-0.024}{-0.095}{0.038} & -0.014 & 2/4 \\
    GIFT-Eval Medium & SF7 & Chronos-2 & 6 & \suppdataci{0.005}{-0.048}{0.048} & 0.011 & 4/6 \\
    GIFT-Eval Medium & SF7 & Toto 2.0 & 6 & \suppdataci{-0.039}{-0.093}{0.002} & -0.016 & 1/6 \\
    GIFT-Eval Medium & SF8 & Chronos-2 & 4 & \suppdataci{0.011}{-0.002}{0.033} & 0.001 & 2/4 \\
    \addlinespace[2pt]
    GIFT-Eval Long & SF6 & Chronos-2 & 4 & \suppdataci{0.034}{-0.064}{0.153} & 0.011 & 3/4 \\
    GIFT-Eval Long & SF6 & Toto 2.0 & 4 & \suppdataci{-0.012}{-0.029}{0.005} & -0.012 & 1/4 \\
    GIFT-Eval Long & SF7 & Chronos-2 & 6 & \suppdataci{-0.006}{-0.092}{0.082} & -0.014 & 2/6 \\
    GIFT-Eval Long & SF7 & Toto 2.0 & 6 & \suppdataci{0.195}{-0.034}{0.629} & -0.008 & 3/6 \\
    GIFT-Eval Long & SF8 & Chronos-2 & 3 & \suppdataci{-0.005}{-0.010}{-0.002} & -0.003 & 0/3 \\
    \addlinespace[2pt]
    fev-bench Mini-20 & SF6 & Chronos-2 & 8 & \suppdataci{0.048}{-0.005}{0.128} & 0.003 & 6/8 \\
    fev-bench Mini-20 & SF6 & TiREx-2 & 8 & \suppdataci{-0.020}{-0.056}{0.001} & -0.003 & 2/8 \\
    fev-bench Mini-20 & SF6 & Toto 2.0 & 8 & \suppdataci{0.135}{-0.010}{0.402} & -0.001 & 4/8 \\
    fev-bench Mini-20 & SF7 & Chronos-2 & 8 & \suppdataci{0.013}{0.004}{0.024} & 0.009 & 7/8 \\
    fev-bench Mini-20 & SF7 & TiREx-2 & 8 & \suppdataci{-0.010}{-0.029}{0.003} & 0.001 & 5/8 \\
    fev-bench Mini-20 & SF7 & Toto 2.0 & 8 & \suppdataci{0.505}{0.002}{1.495} & 0.014 & 7/8 \\
    fev-bench Mini-20 & SF8 & Chronos-2 & 8 & \suppdataci{2.935}{-0.299}{8.982} & 0.150 & 5/8 \\
    fev-bench Mini-20 & SF8 & TimesFM 2.5 & 6 & \suppdataci{0.037}{0.004}{0.079} & 0.028 & 5/6 \\
    fev-bench Mini-20 & SF8 & TiREx-2 & 8 & \suppdataci{0.038}{-0.530}{0.706} & 0.004 & 5/8 \\
    \bottomrule
  \end{tabular}
\end{table}

\begin{table}[h]
  \centering
  \caption{Complete single-realization Chronos-2 trajectories for official quantile-loss and Forecast-Change NRMSE objectives. Values use the panel aggregation on training seed 2026092406 and the same 11,269 samples. This block reports one training seed, so repeated-seed uncertainty is a separate analysis.}
  \label{tab:app-complete-finetuning-curves}
  \scriptsize
  \setlength{\tabcolsep}{5.0pt}
  \begin{tabular}{@{}rrrrr@{}}
    \toprule
    Step & Official MASE & Official FC-NRMSE & FC-NRMSE-objective MASE & FC-NRMSE-objective \\
    \midrule
    0 & 0.725 & 0.641 & 0.725 & 0.641 \\
    1k & 0.681 & 0.593 & 0.778 & 0.533 \\
    2k & 0.665 & 0.556 & 0.804 & 0.488 \\
    3k & 0.655 & 0.582 & 0.816 & 0.473 \\
    4k & 0.645 & 0.587 & 0.815 & 0.483 \\
    5k & 0.641 & 0.606 & 0.814 & 0.477 \\
    6k & 0.635 & 0.605 & 0.819 & 0.475 \\
    7k & 0.632 & 0.607 & 0.820 & 0.476 \\
    8k & 0.630 & 0.614 & 0.819 & 0.478 \\
    9k & 0.629 & 0.616 & 0.820 & 0.471 \\
    10k & 0.628 & 0.615 & 0.820 & 0.473 \\
    \bottomrule
  \end{tabular}
\end{table}


\subsection{Evaluation Coverage and Scope}
\label{app:evaluation-coverage}
\label{app:evidence-identity}

\subsubsection{Evaluation settings}

Table~\ref{tab:app-evaluation-settings} distinguishes the main panels from the supplementary analyses and records their support.

\begin{table}[t]
  \centering
  \caption{Evaluation panels and their declared support. Detailed source provenance and run identities are retained with the analysis record.}
  \label{tab:app-evaluation-settings}
  \scriptsize
  \setlength{\tabcolsep}{4.0pt}
  \begin{tabularx}{\textwidth}{@{}p{0.18\textwidth}p{0.27\textwidth}X p{0.20\textwidth}@{}}
    \toprule
    Panel & Evaluation setting & Main support & Provenance \\
    \midrule
    GIFT-Eval Short latest & Frozen latest Short panel & 55 configurations; 3 unordered variants; 11 models & Analysis record \\
    GIFT-Eval Medium latest & Frozen latest Medium panel & 21 configurations; 10 models & Analysis record \\
    GIFT-Eval Long latest & Frozen latest Long panel & 21 configurations; 10 models & Analysis record \\
    fev-bench Mini-20 latest & Frozen latest Mini-20 panel & 20 configurations; 11 models & Analysis record \\
    Input removal & Matched full-input and target-only views & Same treatment contracts; assessed target history and future fixed & Separate block \\
    Stability (latest) & Five augmentation seeds on Short & 55 configurations; 7 models & Analysis record \\
    Stability (earlier) & Fixed ten-task Short panel & Ten augmentation seeds & Historical block \\
    Fine-tuning & One shared generated corpus and two objectives & 11,269 samples; 10,000 steps; checkpoints every 1,000 steps & Single realization \\
    Model shortlisting & Frozen window-level score cache & 15,673 common-scored windows; 55 configurations; 19 source families; 8-model roster & CPU reconstruction \\
    \bottomrule
  \end{tabularx}
\end{table}

The model-shortlisting analysis uses a frozen window-level score cache and history-profile strengths. Its reconstruction record contains the source-group mapping, analysis procedure, results, reference checks, and redacted provenance; the raw time series, point forecasts, model weights, and private inference endpoints are outside this compact record.

The small-model paired-ranking analysis separates training, generation, evaluation, and analysis inputs. The compact handoff contains stage-level summaries; a full end-to-end replay additionally requires the model checkpoints, full predictions, and source time series.

\subsubsection{Summary of evidence}

Table~\ref{tab:app-evidence-summary} connects each claim family to the corresponding evaluation evidence and presentation in the paper.

\begin{table}[t]
  \centering
  \caption{Summary of evaluation evidence used for the current results.}
  \label{tab:app-evidence-summary}
  \scriptsize
  \setlength{\tabcolsep}{4.0pt}
  \begin{tabularx}{\textwidth}{@{}p{0.23\textwidth}X X@{}}
    \toprule
    Claim family & Evaluation evidence & Result presentation \\
    \midrule
    SF1--SF8 interventions & Structural constructions and their ordered control settings & Feature-level response profiles and heatmaps \\
    Generated-sample accuracy and response & Four evaluation panels with suite-specific support & MASE and Forecast-Change NRMSE tables \\
    Structural-response profiles & Feature-local comparisons on the eligible model support & Main and supplementary heatmaps \\
    Refresh stability & Five-seed and earlier fixed-panel comparisons & Stability summary and figure \\
    Auxiliary-input utility & Matched full-input and target-only comparisons & Auxiliary-Input Gap table and figure \\
    Training-objective trade-off & Two objectives trained on the same sample set & Fine-tuning curves and objective comparison \\
    \bottomrule
  \end{tabularx}
\end{table}

\subsubsection{Evaluation inventory and analysis coverage}

Table~\ref{tab:app-current-execution-inventory} gives the evaluation inventory for the four main panels. Candidates and References count evaluated windows, generated samples count admitted materializations, and predictions count model forecasts. failure counts refer to issued requests, and metric-specific inapplicable cases remain separate from scored cases.

\begin{table}[t]
  \centering
  \caption{Evaluation inventory across the four main panels.}
  \label{tab:app-current-execution-inventory}
  \scriptsize
  \setlength{\tabcolsep}{5.0pt}
  \resizebox{\columnwidth}{!}{%
  \begin{tabular}{@{}lrrrrrrr@{}}
    \toprule
    Suite & Configs & Candidates/References & Generated samples & Models & Predictions & Failures & Horizon \\
    \midrule
    GIFT-Eval Short & 55 & 15,745 & 17,773 & 11 & 399,740 & 0 & 30--60 \\
    GIFT-Eval Medium & 21 & 4,259 & 6,875 & 10 & 127,380 & 0 & 480--600 \\
    GIFT-Eval Long & 21 & 4,179 & 6,552 & 10 & 121,810 & 0 & 720--900 \\
    fev-bench Mini-20 & 20 & 2,809 & 5,609 & 11 & 121,902 & 0 & 5--288 \\
    \bottomrule
  \end{tabular}}
\end{table}

    The four latest panel summaries contribute 88, 80, 80, and 88 model--feature cells for Short, Medium, Long, and FEV, respectively, with the filtered prediction counts shown above. The stability block contains 35 overall seed--model rows in the seven-model view, with feature-level rows retained in the compact summaries. The input-removal block preserves task-level and coverage summaries under the retained model roster. The fine-tuning block contains 22 curve rows and two objective summaries, all from one training realization and a shared 11,269-sample corpus. These counts index downstream records: panel configuration is the latest reduction unit, seed indexes the observed stability panel, and the fine-tuning curves share one training realization. The analysis encodes unavailable, unsupported, not-run, failed, invalid, and metric-undefined states as nonnumeric and preserves their denominators; NRMSE additionally distinguishes scoreable instances from low-truth-effect and no-observed-effect states as defined in Appendix~\ref{app:protocol}.

\FloatBarrier

\subsection{Reproduction Levels and Procedures}
\label{app:reproduction-guide}

This subsection defines three reproducibility levels: checking the consistency of reported summaries, reconstructing the tables and figures from compact numerical summaries, and repeating the complete data construction and forecasting evaluation. The first two levels are the reconstruction scope described by the compact records; the third states the inputs required for a fresh end-to-end run.

\subsubsection{Reproduction levels}

\begin{table}[h]
  \centering
  \caption{Reproduction levels and resource requirements.}
  \label{tab:app-reproduction-levels}
  \scriptsize
  \setlength{\tabcolsep}{4.5pt}
  \begin{tabularx}{\textwidth}{@{}p{0.20\textwidth}p{0.28\textwidth}X@{}}
    \toprule
    Level & Declared environment & Purpose \\
    \midrule
    Summary consistency & Linux or macOS; CPU; less than 1~GB memory and disk & Checks row counts, dimensions, and numerical agreement for representative tables and figures \\
    Table and figure reconstruction & Linux or macOS; Python 3.11 or later; 4 CPU cores, 8~GB RAM, and 5~GB free disk recommended & Recomputes the four analysis blocks reported in the paper from compact numerical summaries \\
    End-to-end evaluation & Linux x86-64; Python 3.12; CUDA 12.8-compatible NVIDIA driver; 32~GB VRAM per worker; 16 CPU cores and 64~GB RAM recommended & Defines the environment for reconstructing treatments, evaluating the models, and recomputing the reported analyses \\
    \bottomrule
  \end{tabularx}
\end{table}

\subsubsection{Reconstruction procedure}

Summary consistency checks compare the numerical summaries with the displayed tables and figures. Table and figure reconstruction recomputes the four main evaluation summaries, the auxiliary-input comparison, the stability analysis, and the fine-tuning curves from compact numerical inputs. End-to-end evaluation is the broader procedure: it requires the benchmark data, model checkpoints, full predictions, and computational resources needed to repeat profiling, augmentation, inspection, forecasting, and metric aggregation.

The model-shortlisting package provides a CPU-only reconstruction route from the frozen window-level score cache. It rebuilds the source splits, leave-family rankings, pooled sensitivity grid, tables, figures, and verification without CaFE, GPU hardware, model weights, raw time series, or network access. Its numerical checks reproduce the frozen package outputs for the declared reconstruction scope.

\subsubsection{Chronos-2 objective comparison}

The Chronos-2 comparison uses the same 11,269-sample corpus for the standard quantile-loss and response-focused objectives. Both branches use the same learning rate, model modules, training seed, schedule, context lengths, and evaluation conditions, with checkpoints every 1,000 steps through 10,000 steps. The comparison therefore isolates the change in training objective for this realization.

\FloatBarrier

\end{document}